%% file: TAI_template.tex
\documentclass[journal]{IEEEtai}

\usepackage[colorlinks,urlcolor=blue,linkcolor=blue,citecolor=blue]{hyperref}
\usepackage{amsmath} 
\usepackage{amssymb}
\usepackage{amsfonts}
\usepackage{cleveref}
\usepackage{algpseudocode,algorithm}
\algnewcommand\RETURN{\State \algorithmicreturn}

\algnewcommand{\LineComment}[1]{\State \(\triangleright\) \textit{\textit{#1}}}
\usepackage{xcolor}
\definecolor{green}{rgb}{0.0, 0.5, 0.0}
\usepackage{booktabs}
\usepackage{graphicx}
\usepackage{pgfplots}
\usepackage{tikz}
\pgfplotsset{compat=newest}
\usepgfplotslibrary{groupplots}
\usepackage[normalem]{ulem}
\usepackage{tikz}
\usetikzlibrary{arrows,matrix,positioning}
\usepackage[normalem]{ulem}  

\begin{document}

\title{Active Learning Using Aggregated Acquisition Functions: Accuracy and Sustainability Analysis}

\author{Cédric Jung, Shirin Salehi, \IEEEmembership{Member, IEEE}, and Anke Schmeink, \IEEEmembership{Senior Member, IEEE}

\thanks{This work was supported by the Federal
Ministry of Education and Research (BMBF, Germany) as part of NeuroSys:
Impulse durch Anwendungen (Projekt D) under Grant 03ZU1106DA.}
\thanks{S. Salehi and A. Schmeink are with the Chair of Information Theory and Data Analytics (INDA), RWTH Aachen University, 52074 Aachen, Germany (e-mail: shirin.salehi@inda.rwth-aachen.de; anke.schmeink@inda.rwth-aachen.de)}
\thanks{C. Jung was with the Chair of Information Theory and Data Analytics (INDA), RWTH Aachen University, 52074 Aachen, Germany. He is now with the Automation and Control
Institute, Technische Universität Wien (TUW), 1040 Vienna, Austria, and also
with the AIT Austrian Institute of Technology GmbH, 1210 Vienna, Austria
(e-mail: jung@acin.tuwien.ac.at)}
}


\maketitle

\begin{abstract}
Active learning (AL) is a machine learning (ML) approach that strategically selects the most informative samples for annotation during training, aiming to minimize annotation costs. This strategy not only reduces labeling expenses but also results in energy savings during neural network training, thereby enhancing both data and energy efficiency. In this paper, we implement and evaluate various state-of-the-art acquisition functions, analyzing their accuracy and computational costs, while discussing the advantages and disadvantages of each method. Our findings reveal that representativity-based acquisition functions effectively explore the dataset but do not prioritize boundary decisions, whereas uncertainty-based acquisition functions focus on refining boundary decisions already identified by the neural network. This trade-off is known as the exploration-exploitation dilemma.
To address this dilemma, we introduce six aggregation structures: series, parallel, hybrid, adaptive feedback, random exploration, and annealing exploration. Our aggregated acquisition functions alleviate common AL pathologies such as batch mode inefficiency and the cold start problem. Additionally, we focus on balancing accuracy and energy consumption, contributing to the development of more sustainable, energy-aware artificial intelligence (AI).
We evaluate our proposed structures on various models and datasets. Our results demonstrate the potential of these structures to reduce computational costs while maintaining or even improving accuracy. Innovative aggregation approaches, such as alternating between acquisition functions such as BALD and BADGE, have shown robust results. Sequentially running functions like $K$-Centers followed by BALD has achieved the same performance goals with up to 12\% fewer samples, while reducing the acquisition cost by almost half.

\end{abstract}


\begin{IEEEkeywords}
Active learning, data-efficiency, energy efficiency, aggregated acquisition functions, exploration-exploitation dilemma, sustainability.
\end{IEEEkeywords}

\section{Introduction}


\IEEEPARstart{T}{he} active learning (AL) framework in machine learning (ML) involves actively selecting and labeling training data to enhance model performance. The main objective is to reduce costs associated with obtaining labeled data by strategically choosing the most informative instances for annotation. In today's machine learning landscape, active learning stands out as particularly pertinent, especially in cases where there is an abundance of unlabeled data that is easily accessible while the labeling process is difficult, time-consuming, or expensive~\cite{ren2021survey}.

In recent years, the field of artificial intelligence (AI) has seen remarkable progress due to the exponential increase in computational power and the availability of large-scale datasets. This progress has resulted in the development of complex models and innovative applications. However, these models require substantial amounts of energy, leading to increased greenhouse gas emissions and a larger carbon footprint within the AI industry~\cite{strubell2019energy, Amodei}. To address these issues, the concept of green AI has emerged~\cite{schwartz2020green}, emphasizing energy efficiency and sustainability alongside traditional metrics like accuracy. Furthermore, the training of these models necessitates processing vast amounts of data~\cite{rydning2018digitization}, which in turn consumes substantial energy and contributes significantly to the carbon footprint. AL shows promising potential to enhance both data and energy efficiency~\cite{salehi2023data, salehi2023active, xu2021survey}. 

\begin{figure}
\begin{center}
\includegraphics[width=0.45\textwidth, keepaspectratio]{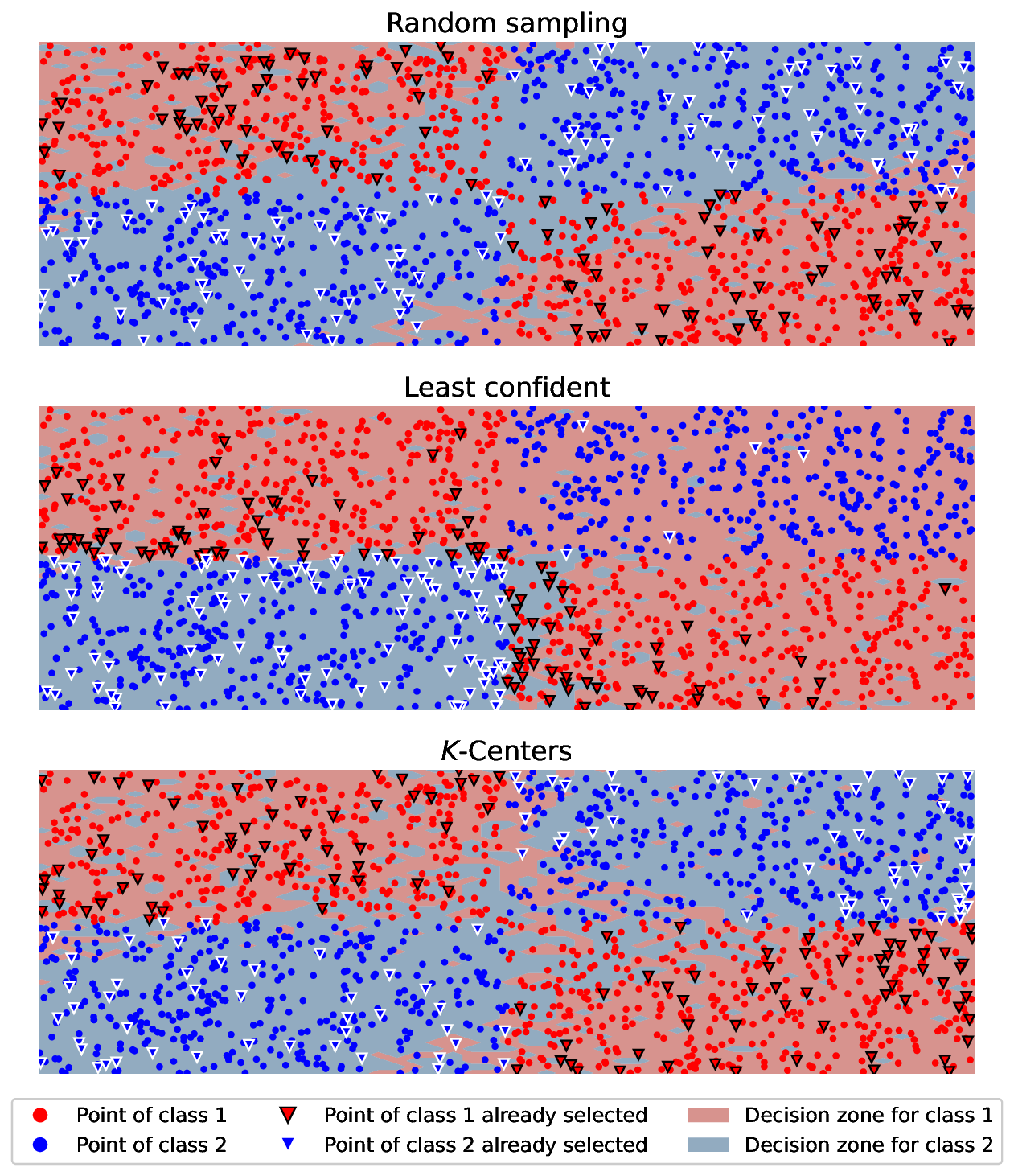}
\caption{Illustration of selected samples and current decision boundary of grid toy dataset at acquisition step 20 with a budget $b = 10$ using a simple fully connected network (FCN) with the following structure: dense-relu-dropout-dense-relu-dense-softmax.}
\label{fig:exploration_exploitation_toyexample}
\end{center}
\vspace{-5mm}
\end{figure}
In active learning, an acquisition function is employed to pick out samples from a vast pool of unlabeled data.
This decision is guided by the current iteration of the neural network, which has been trained with samples selected during previous acquisition rounds.
Diverse acquisition functions have been developed following different paradigms or implementation principles. 
Each of these approaches has its own set of strengths and weaknesses, exhibiting varied behavior depending on the datasets and model architecture utilized. 
Consequently, one function may over-perform the other, or not, depending on these factors.
Therefore, there is no one-fits-all solution for AL.

In AL one wants to minimize the number of data to request. The challenge lies in choosing between exploring new, potentially informative data points and exploiting the current model's knowledge to improve performance. This is called the \textit{exploration-exploitation dilemma}~\cite{yin_similarity_2017, islam_dynamic_exploration_exploitation_2024, jung2024active}. Exploitation involves selecting samples that the current model is uncertain about or likely to be misclassified, aiming to refine the model's decision boundaries in areas of uncertainty. In contrast, exploration involves choosing samples that are different from those already labeled, potentially introducing new concepts or classes that the model has not encountered before. 
In summary, the exploration strategy aims to cover regions of the input space, while the exploitation strategy looks to refine the boundary decision.

Balancing exploration and exploitation in AL is crucial to maximize the efficiency of the labeling process. Over-reliance on exploitation may lead to the model being stuck in local optima or missing out on valuable insights from unexplored regions of the data space. In contrast, excessive exploration can result in labeling redundant or less informative data points, wasting annotation resources.
The exploration-exploitation dilemma is illustrated in Fig. \ref{fig:exploration_exploitation_toyexample}. In this figure, the grid toy dataset is used to show the selection of samples according to random sampling, least confident, and $K$-Centers, respectively. The decision zones depict the results of the model evaluation trained on the samples selected for 20 acquisition rounds with a budget of $b = 10$. One can clearly see that least confident, an uncertainty-based acquisition function, is selecting samples at the boundary decisions of the bottom-left part. However, it fails to discover the two other boundary decisions, resulting in poor accuracy with such a simple problem (77\% of accuracy). On the other hand, the coreset-inspired acquisition function $K$-Centers succeeds in covering the dataset, but has a fuzzy boundary decision and achieves higher performance (94\% of accuracy) compared to random sampling (86\% of accuracy). With this example, one can clearly see the advantages and disadvantages of acquisition paradigms.


In order to improve the performance of AL acquisition functions, one can consider combining different acquisition functions instead of conceiving other sophisticated acquisition functions that can only be effective in some edge cases.
Intuitively, the combination of methods can enable us to leverage the advantages of some methods while mitigating the limitations of others. 
In particular, the contributions of this paper can be summarized as follows:

\begin{itemize}
    \item Evaluate basic and state of the art baseline acquisition functions in terms of accuracy and energy consumption.
   \item Proposing various structures to aggregate multiple acquisition functions, aiming to enhance accuracy while keeping computational costs moderate by strategically combining acquisition functions with different characteristics.
    \item Assess the performance of the proposed approaches across a variety of model structures and datasets, in order to evaluate the generalization and robustness of the proposed approaches.
\end{itemize}

The rest of this paper is organized as follows: We begin by reviewing related works in Section II, providing context and background information on active learning. In Section III, we delve into previous works on aggregating acquisition functions. Section IV presents more detail on proposed structures for aggregating acquisition functions. Finally, these structures are evaluated in Section V, showing the practical applications of such structures, before concluding the paper in Section VI.

\section{Active Learning}

\subsection{Problem settings}

Active learning, also referred to as query learning, strives to enhance a model's performance by minimizing the number of labeled samples needed. Its core concept is straightforward: Different samples in a dataset carry varying degrees of importance for updating the current model, allowing us to pinpoint the most influential examples for training. For effective scaling of machine learning to handle larger problems or more frequent use, it is essential to grasp theoretically and measure the significance of different data points during training and optimization. This enables us to discern valuable samples that contribute meaningfully to the learning process while excluding redundant or uninformative ones confidently. 

AL typically starts with a small training set to train the model. The model then predicts outcomes for unlabeled data, selecting samples that have the greatest potential to improve the model’s accuracy are selected for manual labeling. Labeled samples are added to the training set, and this process repeats until the desired accuracy or labeling budget is reached. 

In the case of pool-based AL and considering the classification tasks, we can define the AL problem as follows. The whole dataset at first presents a small labeled dataset part named $\mathcal{D}_{\mathcal{L}} = \{(x_j, y_j)\}_{j=1}^M$ and a larger unlabeled dataset part named $\mathcal{D}_{\mathcal{U}} = \{x_i\}_{i=1}^N$ where $M \ll N$, $y_i \in \{0, 1\}$ is the class label of $x_i$ for binary classification, or $y_i \in \{1, \ldots, C\}$ for multi-class classification. The process involves selecting instances from the unlabeled dataset $\mathcal{D}_{\mathcal{U}}$ in a greedy manner, guided by a set of informativeness metrics. 
These metrics are encapsulated by what is known as an ``acquisition function". The acquisition function helps to determine the most informative or valuable data points for labeling, facilitating an efficient and effective AL process. 
Thus, in each iteration $t$, we select a sample from $\mathcal{D}_{\mathcal{U}}$ based on the learned model $\mathcal{M}$ and an acquisition function $\alpha(x, \mathcal{M})$, and query their labels from the oracle. Data samples can be selected according to their acquisition score by $x^* = \underset{x \in \mathcal{D}_{\mathcal{U}}}{\operatorname{argmax}}~ \alpha(x, \mathcal{M})$. In general to reduce computational costs, a pool $\mathcal{D}_{\text{pool}}$ is drawn from the unlabeled dataset $\mathcal{D}_{\mathcal{U}}$, on which the acquisition function will be computed: $x^* = \underset{x \in \mathcal{D}_{\text{pool}}}{\operatorname{argmax}}~ \alpha(x, \mathcal{M})$. 


\subsection{Batch mode active learning (BMAL)}


Although historical methods acquire one sample at a time \cite{settles_active_2009}, \cite{gal_deep_2017}, \cite{houlsby_bayesian_2011}, since deep neural networks (DNNs) are computationally heavy, training a new model with a single training sample is highly impractical. Therefore, methods are developed in order to acquire multiple training samples at a time. Moreover, even if it was computationally feasible, a single sample would not have a significant impact statistically on the model given the optimization process.
Therefore, in batch mode active learning (BMAL), in each iteration $t$, we select a batch of samples $\mathcal{D}_t$ with batch size $b = |\mathcal{D}_t|$ from $\mathcal{D}_{\mathcal{U}}$ instead of only one sample. Data samples can be selected according to their acquisition score by $\mathcal{D}_t^* = {\overset{b}{\operatorname{argmax}}}_{x \in \mathcal{D}_{\text{pool}}} \alpha(x, \mathcal{M})$, where the superscript $b$ indicates selection of the top $b$ points. This is known as \textit{Top-$K$} in the literature. However, this can be inefficient if multiple similar samples are selected, leading to redundant information. Acquisition function designs must typically take into account typically three criteria: \textit{informativeness}, \textit{representativeness}, and \textit{diversity}. 

\begin{enumerate}
    \item \textit{Informativeness} or \textit{Uncertainty} is a function $\tau(x)$ that quantifies how much a classifier is expected to benefit from selecting a specific sample $x$. There are several categories of informativeness functions. One can divide such functions into data-based informativeness, which only relies on $\mathcal{D}_{\mathcal{L}}$ and $\mathcal{D}_{\mathcal{U}}$, and model-based informativeness, which only relies on the trained classifier. 
    \item \textit{Diversity} is a function $div(B)$ that only makes sense when applied to a batch $B = \{x_i\}_{i=1}^b$, contrary to the othermentioned criteria. Intuitively, the batch is the most diverse when the samples are the most dissimilar. From an AL perspective having a high diversity batch is good as this will prevent having huge information overlap. Diversity can be achieved explicitly or implicitly. 
    \item \textit{Representativeness} or \textit{representativity} is a function $rep(x)$ that quantifies how well a sample represents the data set. However, there are multiple interpretations and implementations of representativeness. One can consider the density of samples which can be determined by $k$-NN-density measure as proposed by \cite{zhu_active_2008}, where the density of a batch is given by the average distance between the $k$ most similar samples within $\mathcal{D}_{\text{pool}}$.
\end{enumerate}

\subsection{Aquisition functions: an overview}
As mentioned earlier, an acquisition function or query strategy is utilized by the model to evaluate its uncertainty or to pinpoint instances that are likely to yield the most valuable information. Several query selection strategies are introduced below.
\subsubsection{Uncertainty sampling}

The uncertainty sampling approach is the most frequently used paradigm \cite{settles_active_2009}. An acquisition function based on this framework queries the data samples about which the model $\mathcal{M}$ is most uncertain, given the conditional distribution $P(y|\boldsymbol{x}, \omega)$, where $\omega$ denotes the model parameters. 
This kind of acquisition strategy leads to selecting samples near the decision boundaries. A widely adopted uncertainty sampling strategy,  \textit{Max Entropy} or more concisely \textit{Entropy}, employs Shannon entropy \cite{shannon_mathematical_1948}, as a measure of uncertainty:
\begin{equation}
    \begin{aligned}
    x^*_{\mathbb{H}} &= \underset{x \in \mathcal{D}_{\text{pool}}}{\operatorname{argmax}} -\sum_i P(y_i|x, \omega) \log P(y_i|x, \omega) \\
    &= \underset{x \in \mathcal{D}_{\text{pool}}}{\operatorname{argmax}}~{\mathbb{H}(y|x, \omega)},
    \end{aligned}
\end{equation}
where $\mathbb{H}$ is the entropy function. Simpler uncertainty acquisition functions are also used, namely \textit{Least confident}, also named \textit{Variation ratio}, which selects the samples with the least confidence: 
\begin{equation}
    x^*_{LC} = \underset{x \in \mathcal{D}_{\text{pool}}}{\operatorname{argmax}}~1-P(\hat{y}|x, \omega),
\end{equation}
where $\hat{y} = \underset{y}{\operatorname{argmax}}~P(y|x, \omega)$ indicates the class label with the highest posterior probability under the model $\mathcal{M}$ with its associated parameters $\omega$. Another acquisition function, \textit{Margin sampling}, takes into account the entire label distribution rather than just the most probable label: 
\begin{equation}
    x^*_{MS} =\underset{x \in \mathcal{D}_{\text{pool}}}{\operatorname{argmin}}~P(\hat{y_1}|x, \omega)-P(\hat{y_2}|x, \omega),
\end{equation}
where $\hat{y_1}$ and $\hat{y_2}$ represent the top two class labels with the highest probabilities predicted by the model, respectively. Another strategy, \textit{Mean STD} \cite{kampffmeyer_semantic_2016}, selects the samples that maximize the average standard deviation over all possible classes:
\begin{equation}
    x^*_{Mean STD}=\underset{x \in \mathcal{D}_{\text{pool}}}{\operatorname{argmax}}~\frac{1}{C}\sum_i \sigma_i,
\end{equation}

\begin{equation}
    \sigma_i=\sqrt{\mathbb{E}_{q(\omega)}[P(y_i|x, \omega)^2]-\mathbb{E}_{q(\omega)}[P(y_i|x, \omega)]^2 },
\end{equation}
where $C$ is the number of classes and $q(\omega)$ is the dropout distribution.


\subsubsection{BALD} 



\textit{Bayesian active learning by disagreement} (BALD) \cite{houlsby_bayesian_2011} defines an acquisition function based on epistemic uncertainty. Through BALD the samples are selected in such a way that produces disagreeing prediction with high certainty, \textit{i.e.} stochastic forward passes would have the highest probability to assign the samples to different classes. Specifically, it uses the \textit{mutual information} (MI) between the unknown output and model parameters as a measure of disagreement:
\begin{equation}
    \mathbb{I}(y; \omega | x, \mathcal{D}_{\mathcal{L}}) = \mathbb{H}(y | x, \mathcal{D}_{\mathcal{L}}) - \mathbb{E}_{\omega \sim p(\omega|\mathcal{D}_{\mathcal{L}})} [\mathbb{H}(y | x, \omega)].
\end{equation}

BALD tries to maximize the mutual information or more explicitly  the information gained about the model parameters,  $\mathbb{I}(y; \omega | x, \mathcal{D}_{\mathcal{L}})$, between predictions and model posterior: 
\begin{equation}
    x^*_{BALD} =\underset{x \in \mathcal{D}_{\text{pool}}}{\operatorname{argmax}}~\mathbb{I}(y; \omega | x, \mathcal{D}_{\mathcal{L}}).
\end{equation}

A computationally cheaper way to avoid computing the mutual information gain was found by \cite{kirsch_stochastic_2023}, while it is still capable of mitigating information overlap. The authors introduce a weighted importance sampling across $\mathcal{D}_{\text{pool}}$ taking the individual scores given by BALD as input. They introduce a class of acquisition functions called \textit{stochastic acquisition function} which includes \textit{PowerBALD}. 

\subsubsection{Greedy $K$-Centers} 

Introduced in \cite{sener_active_2018}, \textit{Greedy $K$-Centers} or simply \textit{$K$-Center}s is a core-set inspired acquisition function which allows to enforce diversity of selected samples on the unlabeled batch. The core-set selection problem is to select a subset of the whole dataset so that the model trained on this subset performs as closely as possible to the model trained on the entire dataset. This method does not include the notion of informativeness directly, but instead selects instances that best cover the dataset by maximizing the minimal distance to the labeled set $\mathcal{D}_{\mathcal{L}}$ on the learned feature space (last hidden layer of the neural network). 


\subsubsection{BADGE} 

The state-of-the-art acquisition function \textit{Batch Active Learning by Diverse Gradient Embeddings} (BADGE), has been introduced recently in \cite{ash_deep_2020}. It selects samples by first computing the last-layer gradient $g_{x} = \nabla l(x,y=\hat{y}, \omega^{\text{L}})$ (gradient embedding), obtained if the most likely label according to the model, $\hat{y}$, were observed. Here $\omega^{\text{L}}$ refers to the parameters of the last hidden layer. It then uses $k$-Means$++$ to select $b$ samples \cite{arthur_k-means_2007}. The use of $k$-Means$++$ is motivated by the fact that it is a computationally efficient approximation of Determinantal Point Process (DPP) sampling.

The authors who originally proposed BADGE later introduced \textit{Batch Active learning via Information maTrices} (BAIT), a more recent acquisition function that extensively leverages \textit{Fisher Information} (FI)  and is designed for use in a BMAL framework \cite{ash_gone_2021}.
Considering the negative log-likelihood $l(x,y,\omega) = - \log p(y|x;\omega)$, FI can be written as
$\mathcal{I}_{x}(\omega) =   \mathbb{E}_{p(y|x;\omega)} \left[ \nabla_{\omega}^2 l(x,y,\omega)  \right]$. BAIT selects batches by minimizing a Fisher information–based objective:
\begin{equation} \label{eq:fir_bmal}
    \underset{S \subseteq \mathcal{D}_{\text{pool}}, |S| \leq b}{\operatorname{argmin}} \text{tr} \left( \left( \sum_{x \in S} \mathcal{I}_{x}(\omega) \right)^{-1} \mathcal{I}_{\mathcal{U}}(\omega) \right)
\end{equation} with $\mathcal{I}_{\mathcal{U}}(\omega) = \sum_{x \in \mathcal{D}_{\text{pool}}} \mathcal{I}_{x}(\omega)$.
The authors claim that BADGE is in fact an approximation of BAIT, resulting in better result in terms of the number of samples required. BAIT has the advantage over BALD to consider all the possible labels $y$, while BADGE only considers the most probable label. However, BADGE is $C$ times faster than BAIT, where $C$ is the number of classes.

\subsubsection{Submodularity-based acquisition function}

Submodular functions inherently represent concepts of information, diversity, and coverage in various applications \cite{fujishige2005submodular}. Additionally, they can be optimized effectively using remarkably straightforward algorithms. A greedy algorithm \cite{nemhauser_analysis_1978} guarantees that the cardinality constrained submodular maximization problem can be approximated to a factor of $1 - \frac{1}{e}$. It is therefore natural that they have been used in AL \cite{wei_submodularity_2015}. The discussion below will present only two of these functions.
\paragraph{Facility location}
Facility location (FL) function \cite{mirchandani1990discrete} aims to represent the data by identifying a subset of representative elements. This submodular function goal is to have the most representative subset. The idea is similar to $k$-medoids clustering. FL is solved by acquiring in a greedy manner the samples that have the maximum similarity to the groundset:
\begin{equation}
\begin{aligned}
\mathcal{D}_t^* &= \underset{{\mathcal{D}_t \subseteq \mathcal{D}_{\text{pool}}}}{\operatorname{argmax}} ~\alpha_{\text{FL}}(\mathcal{D}_t, \mathcal{D}_{\mathcal{L}}, \mathcal{M}) \\
&= \underset{{\mathcal{D}_t \subseteq \mathcal{D}_{\text{pool}}}}{\operatorname{argmax}} ~ \sum_{x_i \in \mathcal{D}_{\text{pool}}} \max_{x_j \in {\mathcal{D}_t}} s(x_i,x_j)
\end{aligned}
\end{equation}
where $s(x_i,x_j)$ denotes the similarity between $x_i$ and $x_j$.
\paragraph{Disparity min}
In contrast to FL, \textit{disparity} functions are diversity-based functions that strive to acquire a diverse set of samples, with the objective of minimizing similarity among elements in the selected subset by maximizing the minimum distance between pairs.  The minimum disparity function (\textit{disparity min}) \cite{dasgupta2013summarization} can be expressed as follows:
\begin{equation}
\begin{aligned}
\mathcal{D}_t^* &= \underset{{\mathcal{D}_t \subseteq \mathcal{D}_{\text{pool}}}}{\operatorname{argmax}} ~\alpha_{\text{Disparity min}}(\mathcal{D}_t, \mathcal{D}_{\mathcal{L}}, \mathcal{M}) \\
&= \underset{{\mathcal{D}_t \subseteq \mathcal{D}_{\text{pool}}}}{\operatorname{argmax}} ~ \min_{x_i,x_j \in {\mathcal{D}_t}} d(x_i,x_j)
\end{aligned}
\end{equation}
where $d(x_i,x_j)$ denotes the distance between $x_i$ and $x_j$.
This function is not submodular, but according to \cite{dasgupta2013summarization} it can also be optimized via a greedy algorithm. 

\section{Related works}

The concept of aggregating acquisition functions in a series structure, as outlined in this work, is not entirely novel. Englhardt et al. \cite{adrian_englhardt_finding_2020} explored this in a one-class problem context. Similarly, Wei et al. \cite{wei_submodularity_2015} introduced the \textit{Filtered Active Submodular Selection} (FASS) framework. Although they did not explicitly define a series structure, their multistage framework—initially selecting $\kappa b$ samples based on uncertainty scores and the most likely label, followed by the application of a submodular function to select the final $b$ samples—aligns closely with this concept.

The concept of information density weighting, as discussed by Settles et al. \cite{settles_analysis_2008}, integrates diversity and uncertainty. The idea is to counter-weight the uncertainty in the selection of samples to also take into account the most representative points of the data distribution. This means that the acquisition function should also select samples in the dense regions of the data distribution. 
The information density acquisition function is introduced by \cite{settles_analysis_2008} as follows and is composed of two parts:
\begin{equation}
 x^{*}_{\text{ID}} = \underset{x \in \mathcal{D}_{\text{pool}}}{\operatorname{argmax}} ~  \phi(x) \times \left( \frac{1}{|\mathcal{D}_{\text{pool}}|-1} \sum_{z \in \mathcal{D}_{\text{pool}} \setminus \{x\} } \text{sim}(x, z) \right)^\beta,
\end{equation}
where $\phi(x)$ represents the informativeness of $x$ according to a first acquisition function (uncertainty-based for example), while the second term is the average similarity to all other samples from the data distribution $\mathcal{D}_{\text{pool}}$. The hyperparameter $\beta$ allows one to balance the importance of the two terms. This method ensures that uncertain samples with redundant information are not repeatedly selected, thereby improving the efficiency of the active learning process. Authors in~\cite{demir_interactive_2011} also implemented such an approach leveraging support vector machine's (SVM) ability to compute distance.

Incorporating a clustering step, as suggested in \cite{demir_interactive_2011}  and \cite{wei_submodularity_2015}, can enhance the selection of diverse samples, ensuring broad coverage of the feature space.  The pre-clustering approach from Nguyen et al. \cite{nguyen_pre_cluster_2004} further refines this by dynamically adjusting cluster sizes to balance exploration and exploitation.

The diversity measure in uncertainty sampling, as explored by Yin et al. \cite{yin_similarity_2017}, addresses the exploitation-exploration dilemma. They define exploitation as selecting maximally uncertain and minimally redundant instances and exploration as selecting samples most diverse from the labeled set. Their formula $I(S) = E(S) - \beta R(S)$ incorporates redundancy reduction directly into the acquisition process.
Gu et al. \cite{gu_adaptive_active_2015} proposed a similar method using SVM and Gaussian kernel to compute similarity. Although this method does not involve deep learning, it focuses on minimizing redundancy while maximizing information gain.


Fixed combinations of acquisition functions, with predefined trade-off parameter $\beta$, have evolved into adaptive combinations where the trade-off parameter adjusts dynamically. Our feedback-driven method is inspired by this adaptive approach. However, it selects among different acquisition functions based on the feedback signal on acquisition rounds rather than on a fixed ratio over the entire acquisition process.

The feedback-driven acquisition functions have been studied in works such as Ebert and Ralf \cite{ebert_ralf_2012} and Cheng et al. \cite{cheng_feedback-driven_2013}. Their approach weights the contribution of uncertainty-based and diversity-based acquisition functions using a dynamic trade-off ratio $\beta(t)$. They implemented a weighted sum of the ranks provided by two acquisition functions. This is similar to one of our proposed approaches called parallel-ranked.
However, they use a single-sample selection at a time, which contrasts with our BMAL approach. Their approach uses the following equation for the acquisition round $t$:

\begin{equation}
    \alpha_t(x_i,\mathcal{M}) = \beta(t)r(U(x_i)) + (1-\beta(t))r(D(x_i)),
\end{equation}
with $r$ being the ranking function, and $\beta(t)$ a trade-off ratio. A batch can be selected in a Top-$K$ manner:

\begin{equation}
    \mathcal{D}_t^* = {\overset{b}{\operatorname{argmin}}}_{x \in \mathcal{D}_{\text{pool}}} \alpha(x,\mathcal{M})
\end{equation}

Their method thus combines the disadvantages of parallel-ranking and Top-$K$ selection, while our approach provides feedback to choose between acquisition functions rather than a weighted rank combination, potentially reducing Top-$K$ selection pathologies within the combined acquisition functions.


Several studies have employed bandit approaches to balance exploration and exploitation in active learning in the same manner. For instance, Tran et al. \cite{tran_combining_2018} used Thompson sampling, Baram et al. \cite{baram_online_2004} applied the EXP4 algorithm, and Hu et al. \cite{hu_egal_2010} explored similar methodologies. These approaches dynamically adjust the exploration-exploitation trade-off, striving for optimal sample selection under finite data constraints, acknowledging that even the best heuristics may fall short due to poor classifiers trained on limited data. 

 Recent work by Flesca et al. \cite{flesca_meta-active_2024} is a meta-learning approach which computes importance scores based on criteria like direct similarity or leave-one-out distance and trains a regression model on pairs of labeled instances, each associated with an importance score. This approach aims to predict the most likely importance score for future acquisitions, enhancing the decision-making process in active learning. 
A recent work \cite{islam_dynamic_exploration_exploitation_2024} introduced dynamic trade-offs in exploration and exploitation, emphasizing the need for iterative updates to the trade-off parameter as new data is queried. This contrasts with classical approaches that use a fixed trade-off parameter, highlighting the need for flexibility in active learning strategies.



\section{Proposed structures for aggregated acquisition functions}

In this section, the proposed structures for aggregating acquisition functions are introduced, the motivation behind them is explained, and their expected pros and cons are discussed.

\subsection{Parallel and parallel-ranked structures}

In \textit{parallel} and \textit{parallel-ranked} structures, acquisition functions run in parallel, and then the results are combined.

\subsubsection{Parallel-ranked} \label{sec:parallel-ranked}

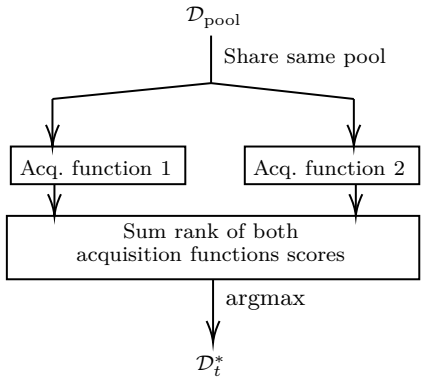
\begin{figure}

\begin{center}

\input{figures/structure/figure_parallel_ranking_small}

\caption{Schema of parallel-ranked combination of two acquisition functions.}
\label{fig:parallel-ranked_structure}
\end{center}
\vspace{-5mm}
\end{figure}

A straightforward approach is to sum the ranks provided by two acquisition functions and then select the Top-$K$ samples, as shown in Fig. \ref{fig:parallel-ranked_structure}. The combined acquisition function can thus be expressed as:
\begin{equation}
\alpha(x_i,\mathcal{M}) = r(\alpha_1(x_i,\mathcal{M})) + r(\alpha_2(x_i,\mathcal{M})),
\end{equation}
with $r$ the ranking function $r:\mathbb{R}\rightarrow\{1,\ldots,u\},~r(F(x_i))=m_i$, where $F(x_i) \leq F(x_j) \Leftrightarrow m_i \geq m_j$ with $m \in \{1, \ldots, u\}$ with $F$ a real-valued function. A batch can be selected in a Top-$K$ manner: 
\begin{equation}
    \mathcal{D}_t^* = {\overset{b}{\operatorname{argmin}}}_{x \in \mathcal{D}_{\text{pool}}} \alpha(x,\mathcal{M})
\end{equation}

This approach, which we call \textit{parallel-ranked}, has two drawbacks.
Firstly, not all the acquisition functions give a score to samples (for example BADGE), which makes this structure infeasible. 
Secondly, scores sometimes cancel each other out. For instance, two acquisition functions can give midscores to a sample, and the sum of them would result in selecting the sample, while none of the acquisition functions considers this sample good enough for selection. This does not give automatically the best of two acquisition functions criteria and can even have lower performance than originally considered acquisition functions alone. Furthermore, it can intensify the pathologies of batch selection. 

\subsubsection{Parallel}

\begin{figure}
\begin{center}
\input{figures/structure/figure_parallel_small}
\caption{Schema of the parallel combination of two acquisition functions. By ``Concat" we refer to the concatenation of the elements of the selected batches into a single batch.}
\label{fig:parallel_structure}
\end{center}
\vspace{-5mm}
\end{figure}
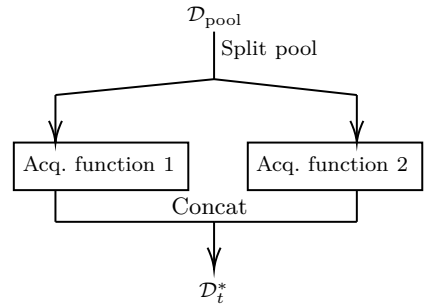

Another approach is to run two acquisition functions in parallel on different subsets of the pool and then dedicate only a part of the budget $b$ to the samples selected via each acquisition function. We call this structure \textit{parallel}. 
This is done because running acquisition functions can be computationally expensive and as we cannot ensure that acquisition functions will not select the same samples, one considers them running only on a part of the pool $\mathcal{D}_\text{pool}$. 
If two acquisition functions are considered to run in parallel, each will select a budget of $\tfrac{1}{2}b$ from a pool of size $\tfrac{1}{2}|\mathcal{D}_\text{pool}|$.
This structure is illustrated in Fig. \ref{fig:parallel_structure}.
Formally, this approach can be written as:
\begin{equation}
\begin{split}
\mathcal{D}_t^* 
  &= \alpha(\mathcal{D}_{\text{pool}}, b, \mathcal{M}) \\
  &= \alpha_{1}\!\left(\mathcal{D}_{\frac{|\mathcal{D}_{\text{pool}}|}{2}}, \tfrac{b}{2}, \mathcal{M}\right) 
     \cup \alpha_{2}\!\left(\mathcal{D}_{\frac{|\mathcal{D}_{\text{pool}}|}{2}}, \tfrac{b}{2}, \mathcal{M}\right).
\end{split}
\end{equation}
Here, the second argument denotes the batch size to be selected, and $\mathcal{D}_{\frac{|\mathcal{D}_{\text{pool}}|}{2}}$ represents a split of the pool $\mathcal{D}_{\text{pool}}$ of cardinality $\tfrac{|\mathcal{D}{\text{pool}}|}{2}$.

\subsection{Series structure} \label{sec:series_structure_intro}

In this approach, the acquisition functions are aggregated in a \textit{series} manner. In the series structure, at least two acquisition functions are applied sequentially, each reducing the pool of data on which the next acquisition function will operate. This structure can be visualized as a sieve with multiple layers.
In this framework, we define a subsample factor $\kappa \geq 1$, which determines the additional budget $\kappa b$ allocated to the first acquisition function.
This extra budget is progressively reduced at the next layer, ultimately converging to the final budget $b$.
Given two acquisition functions $\alpha_1$ and $\alpha_2$, a subsample factor $\kappa \geq 1$ is applied. At acquisition step $t$:
\begin{equation}
    \begin{aligned}
        {\mathcal{D}_{t,1}^*} &= \alpha_{1}({\mathcal{D}_\text{pool}}, \kappa b, \mathcal{M}), \\
        |{\mathcal{D}_{t,1}^*}| &= \kappa b,\\
        {\mathcal{D}_{t}^*} &= \alpha_{2}({\mathcal{D}_{t,1}}, b, \mathcal{M}), \\
        |{\mathcal{D}_{t}^*}| &= b,
    \end{aligned}
\end{equation} where the second argument denotes the batch size to be selected.

This structure has the advantage of reducing the pool for the second acquisition function. Generally, $ |\mathcal{D}_\text{pool}| >> \kappa b$, the cost of using this second acquisition function on top of the first is limited compared to running two acquisition functions on the whole pool. The series structure is detailed in algorithm~\Cref{alg:al-series-structure}.


\begin{algorithm}
\caption{Active learning with series structure}
\label{alg:al-series-structure}
\begin{algorithmic}[1]
\Require Neural network $\mathcal{M} = f(x;\omega)$
\Require Unlabeled dataset $\mathcal{D}_{\mathcal{U}}$
\Require Initial number of samples $M$, 
\Require Total number of iterations $T$
\Require Number of samples in a batch $b$
\Require Acquisition functions list $[\alpha^{(1)}, \alpha^{(2)}, \cdots, \alpha^{(N_{\text{acq}})}]$ containing ${N_{\text{acq}}}$ acquisition functions
\Require Subsample factor list $[\kappa_1, \kappa_2, \cdots, \kappa_{N_{\text{acq}}}]$ containing ${N_{\text{acq}}}$ factors with $\kappa_1 \geq \kappa_2 \geq \cdots \geq \kappa_{N_{\text{acq}}} \geq 1$ 
\State Labeled dataset $\mathcal{D}_{\mathcal{L}} \gets M$ examples drawn randomly from $\mathcal{D}_{\mathcal{U}}$ together with queried labels.
\State Train an initial model $\mathcal{M}_{0}$ on $\mathcal{D}_{\mathcal{L}}$. 
\For{$t = 1, \ldots, T-1$}
    \State Draw $\mathcal{D}_{\text{pool}}$ randomly from $\mathcal{D}_{\mathcal{U}}$
    \State ${\mathcal{D}_t^*}^{(0)} \gets {\mathcal{D}_{\text{pool}}}$
    \For{$i = 1, \ldots, N_{\text{acq}}$}
        \State Select $b$ sample ${\mathcal{D}_t^*}^{(i)} \gets \alpha^{(i)}({\mathcal{D}_t^*}^{(i-1)}, \kappa_i b, \mathcal{M}_{t})$
    \EndFor
    \State ${\mathcal{D}_{\mathcal{U}}} \gets {\mathcal{D}_{\mathcal{U}}} \setminus {{\mathcal{D}_t^*}^{(N_{\text{acq}})}}$
    \State ${\mathcal{D}_{\mathcal{L}}} \gets {\mathcal{D}_{\mathcal{L}}} \cup {{\mathcal{D}_t^*}^{(N_{\text{acq}})}}$
    \State Train a model $\mathcal{M}_{t+1}$ on $\mathcal{D}_{\mathcal{L}}$. 
\EndFor
\State \Return Final model $\mathcal{M}_{T}$.
\end{algorithmic}
\end{algorithm}

\subsection{Hybrid structure}

Another way to combine the acquisition functions is to mix the parallel and series structures. This structure allows us to run two acquisition functions on the same pool like parallel-ranked, while having the knowledge of selected samples by the first one. The only restriction compared to the parallel structure is that one has to execute the functions one after another. This structure is proposed in order to improve the performance of the parallel structure. 
This structure can be written using recursive equations:
\begin{equation}
\begin{aligned}
{\mathcal{D}_{t,0}^*} &= \emptyset, \\
{\mathcal{D}_{t,1}^*} &= \alpha_{1}(\mathcal{D}_{\text{pool}}, b_1, \mathcal{M}), \\
{\mathcal{D}_{t,2}^*} &= \alpha_{2}(\mathcal{D}_{\text{pool}} \setminus {\mathcal{D}_{t,1}^*}, b_2, \mathcal{M}), \\
\mathcal{D}_t^* &= {\mathcal{D}_{t,1}^*} \cup {\mathcal{D}_{t,2}^*}.
\end{aligned}
\end{equation}
with a total budget of $b = b_1 + b_2$. It can be seen that the hybrid structure is just the execution of several acquisition rounds with a total budget $b$ without retraining the model $\mathcal{M}$, which is the costly part.


In the following subsections, we present selection mechanisms for switching between acquisition functions, including adaptive feedback, annealing, and random alternation. The computational cost of employing such a dual-strategy scheme is comparable to that of using a single acquisition function, since only one function is executed in each iteration. The only additional overhead stems from evaluating the selection mechanism itself, which is negligible relative to the overall training cost.

\subsection{Adaptive feedback structure}

A more sophisticated approach selects between acquisition functions based on a feedback signal. To choose the acquisition function $\alpha_{t+1}$ for the next AL iteration $t+1$, we use the loss of the previous model $\mathcal{M}_{t-1}$—not the current model $\mathcal{M}_t$—evaluated on the current selected batch $\mathcal{D}_{t}^*$. Intuitively, this loss is a meaningful metric, as it reflects the model's error on data it has not seen before. We refer to this method as \textit{adaptive feedback} due to its similarity to a feedback control loop.

The key idea is to alternate between exploration-oriented acquisition functions (e.g., $K$-Centers) and exploitation-oriented ones (e.g., BALD) depending on the observed behavior of the feedback metric. Exploration is favored when the model is still discovering new regions of the input space, while exploitation is triggered when uncertainty increases around decision boundaries. This strategy alleviates the cold-start problem in active learning and mitigates the tendency of uncertainty-based functions to focus only on poorly conditioned boundaries. The cost of using such a structure is the same as running acquisition functions alone, as per iteration round, only one acquisition function is running, plus the cost of choosing the feedback. The additional cost, therefore, only depends on the feedback strategy.

In our implementation, the feedback metric is the \emph{previous model loss}:
\begin{equation}
\ell_t = J\!\left(\mathcal{D}_t^*, \omega_{t-1}\right),    
\end{equation}
where $J$ is the loss function and $\omega_{t-1}$ are the parameters of $\mathcal{M}_{t-1}$. Since $\ell_t$ tends to decrease over time due to empirical risk minimization, we normalize it over a sliding window of size $n_\text{window}$:
\begin{equation}
s_t = \frac{\ell_t - \min(\ell_{t-n_\text{window}:t})}{\max(\ell_{t-n_\text{window}:t}) - \min(\ell_{t-n_\text{window}:t})}.    
\end{equation}
Here $s_t \in [0,1]$ serves as a scaled \emph{reward}, where higher values indicate that the selected batch was more challenging for the model.
This idea was inspired by the client selection framework in federated learning proposed by \cite{cho_towards_2022}, where clients are selected based on the larger local loss.

As a trade-off update rule, we maintain a trade-off parameter $\beta_t \in [\epsilon, 1-\epsilon]$ ($\epsilon > 0$) controlling the exploration–exploitation choice:
\begin{equation}
    \beta_t = \max\!\left( \min\!\left( \lambda \beta_{t-1} \exp(s_t), \; 1-\epsilon \right), \; \epsilon \right),    
\end{equation}
with $\beta_{1} = 0.5$ and learning rate $\lambda > 0$.  
The decision rule is:
\begin{equation}
\mathcal{D}_{t}^* =
    \begin{cases}
        \alpha_{\text{explore}}\!\left(\mathcal{D}_{\text{pool}}, b, \mathcal{M}_{t-1}\right), & \text{if } \beta_{t} \leq 0.5, \\
        \alpha_{\text{exploit}}\!\left(\mathcal{D}_{\text{pool}}, b, \mathcal{M}_{t-1}\right), & \text{if } \beta_{t} > 0.5.
    \end{cases}    
\end{equation}
Thus, $\beta_t$ dynamically adjusts the balance between exploration and exploitation based on the feedback metric, increasing the likelihood of exploitation when recent batches have yielded higher loss. The algorithm described above is written in \Cref{alg:al-adaptive-feedback-with-loss}, featuring the integrated adaptive component. 
The choice of $\beta_{1} = 0.5$ is to ensure that the AL process starts with exploring the dataset rather than exploiting biased knowledge.

\begin{algorithm}
\caption{Active learning with adaptive feedback structure using previous model loss on selected batches}
\label{alg:al-adaptive-feedback-with-loss}
\begin{algorithmic}[1]
\Require Neural network $\mathcal{M} = f(x;\omega)$
\Require Unlabeled dataset $\mathcal{D}_{\mathcal{U}}$
\Require Initial number of samples $M$, 
\Require Total number of iterations $T$
\Require Number of samples in a batch $b$
\Require Acquisition functions with exploration $\alpha_\text{exploration}$ and exploitation $\alpha_\text{exploitation}$ paradigms
\Require Size of metric window $n_\text{window}$ and the learning rate of the feedback loop $\lambda$
\State Initialize $\beta_1 = 0.5$ and $\epsilon = 0.1$.
\State Initialize $L$ an empty list of previous model loss on the selected batch.
\State Labeled dataset $\mathcal{D}_{\mathcal{L}} \gets M$ examples drawn randomly from $\mathcal{D}_{\mathcal{U}}$ together with queried labels.
\State Train an initial model $\mathcal{M}_{0}$ on $\mathcal{D}_{\mathcal{L}}$.
\For{$t = 
1, \ldots, 
T
$}
    \LineComment{Selecting acquisition function}
    \If{$\beta_t \leq 0.5$}
        \State Use exploration acquisition function $\alpha_\text{exploration}$.
    \Else
        \State Use exploitation acquisition function $\alpha_\text{exploitation}$.
    \EndIf
    \LineComment{Acquisition of batch}
    \State Draw $\mathcal{D}_{\text{pool}}$ randomly from $\mathcal{D}_{\mathcal{U}}$.
    \State Select $b$ sample $\mathcal{D}_t^* = \alpha(\mathcal{D}_{\text{pool}}, b, \mathcal{M}_{t-1})$.
    \State ${\mathcal{D}_{\mathcal{U}}} \gets {\mathcal{D}_{\mathcal{U}}} \setminus {\mathcal{D}_t^*}$.
    \State ${\mathcal{D}_{\mathcal{L}}} \gets {\mathcal{D}_{\mathcal{L}}} \cup {\mathcal{D}_t^*}$.
    
    \State Train a model $\mathcal{M}_{t}$ on $\mathcal{D}_{\mathcal{L}}$ by getting
    \Statex \hspace{\algorithmicindent} $\omega_{t} = \arg\min_{\omega}
 J(\mathcal{D}_{\mathcal{L}}, \omega)$.

    \LineComment{Computing trade-off parameter}
    \State Add previous model loss on selected batch 
    \Statex \hspace{\algorithmicindent}  $J(\mathcal{D}_t^*, \omega_{t-1})$ to $L$.
    \State Extract previous model loss subset
    \Statex \hspace{\algorithmicindent} $L' = L[{t-{n_\text{window}}}; {t}]$.
    \State Compute moving average on subset $L' = MA(L')$.
    \State Compute scaled score : $s_{t} = \frac{L'[-1] - \min(L')}{\max(L') - \min(L')}$.
    \State Compute trade-off score :
    \State $\beta_t = \max\left(\min\left(\lambda \beta_{t-1}\exp(s_{t}),1-\epsilon\right),\epsilon\right)$.
\EndFor
\State \Return Final model $\mathcal{M}_{T}$.
\end{algorithmic}
\end{algorithm}

\subsection{Annealing exploration structure}

Another way to address the exploration-exploitation dilemma is to implement a selection mechanism with \textit{annealing} behavior. This is inspired by the concept of cosine annealing learning rate in machine learning \cite{loshchilov2017sgdr}. 
In this approach, exploration-oriented acquisition functions are applied exclusively during an initial phase of $T_{\text{initial exploration}}$ AL iterations. After this, the process alternates between exploitation and exploration phases. The first exploitation phase lasts ${T_{\text{exploit}}}_1$ iterations, followed by an exploration phase of fixed length $T_{\text{explore}}$. In subsequent cycles $i \geq 1$, the exploitation phase duration increases multiplicatively according to a rate $r > 1$, i.e.,
${T_{\text{exploit}}}_{i+1} = \left\lfloor r \cdot {T_{\text{exploit}}}_{i} \right\rfloor$.
This results in gradually longer exploitation phases, while exploration phases remain constant in length, thus biasing the process toward exploitation as AL progresses.
Formally, let $\alpha_{\text{explore}}$ and $\alpha_{\text{exploit}}$ denote the respective acquisition functions. At each iteration $t$, the selected batch is given by:
\begin{equation}
\mathcal{D}_t^* =
    \begin{cases}
        \alpha_{\text{explore}}\!\left(\mathcal{D}_{\text{pool}}, b, \mathcal{M}\right), & \text{if $t$ is in an exploration phase}, \\
        \alpha_{\text{exploit}}\!\left(\mathcal{D}_{\text{pool}}, b, \mathcal{M}\right), & \text{if $t$ is in an exploitation phase}.
    \end{cases}    
\end{equation}
The schedule of exploration and exploitation phases is determined by $T_{\text{initial exploration}}$, ${T_{\text{exploit}}}_1$, $T_{\text{explore}}$, and $r$.


\begin{figure}
\input{figures/structure/figure_annealing_small}
\label{fig:annealing_structure_illustration}

\caption{Annealing exploration structure}
\label{fig:annealing_structure}
\vspace{-5mm}
\end{figure}
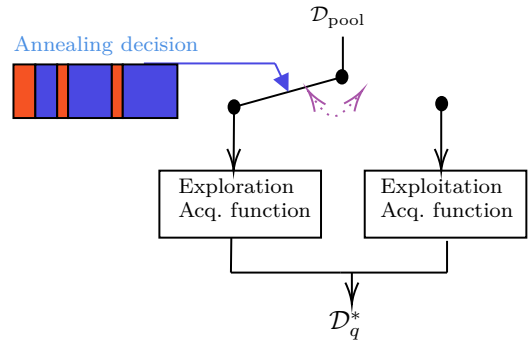

\subsection{Random exploration structure}

In this structure, we simply alternate between two acquisition functions randomly without any kind of feedback from the quality of the selected batch. Formally, let $\{\alpha_{1}, \alpha_{2}\}$ be the two acquisition functions considered. At round $t$, draw $ I_t \sim \text{Uniform}\{1,2\}$ and select $\mathcal{D}_t^* = \alpha_{I_t}\!\left(\mathcal{D}_{\text{pool}}, b, \mathcal{M}\right)$, where $\mathcal{M}$ is the current model and $\mathcal{D}_{\text{pool}}$ is the unlabeled data pool. This approach serves as a baseline to evaluate whether more sophisticated alternation strategies offer measurable improvements.



\section{Experiments}

\subsection{Experimental settings}

\subsubsection{Dataset and models}

\paragraph{CIFAR}
The CIFAR datasets \cite{krizhevsky_cifar-10_2009} are widely used compilations of real-world images, facilitating comparisons in deep learning tasks for computer vision. Named after the Canadian Institute for Advanced Research, these datasets consist of color images representing various real-world objects. The CIFAR datasets have become a standard for establishing baselines in deep learning. 

The CIFAR10 dataset contains 60,000 images, each sized at 32 × 32 pixels and in color. These images are divided into 10 distinct categories, with 6000 images assigned to each category. It is split into a training set of 50,000 images and a test set of 10,000 images. The test set includes 1000 randomly selected images for each category. Similarly, the CIFAR100 dataset comprises 60,000 images, also 32 × 32 pixels in size and in color, but grouped into 100 classes, each with 600 images. Within each class, there are 500 training images and 100 testing images.
It is important to note that the CIFAR100 dataset presents more challenging scenarios compared to CIFAR10. This is primarily due to their increased number of class labels and reduced number of examples per class.
For CIFAR datasets we consider the three following combinations of dataset and model:
\begin{itemize}
    \item CIFAR10 and VGG16 which is VGG with 16 layers
    \item CIFAR10 and ResNet18 which is ResNet with 18 layers 
    \item CIFAR100 and VGG16
\end{itemize}
\paragraph{PTB-XL} We extend our study to the ECG classification task, a time series classification problem essential for detecting cardiovascular diseases. The PTB-XL dataset \cite{wagner_ptb-xl_2020} was selected due to its considerably larger sample size (over 18,000 individuals) compared to ECG5000 \cite{chen_general_2015} (5,000 samples from a single patient). Our experiments focus on the five coarse superclasses of PTB-XL. Following prior work \cite{strodthoff_deep_2021}, we adopt ResNet101, which has demonstrated strong performance on this dataset.

\subsubsection{Implementation details}

We implemented the ResNet, and VGG in Pytorch~\cite{pytorch}. To implement the iterative workflow of AL we have used a version of modAL~\cite{modAL2018}. The implementation of the submodular-based acquisition functions utilizes the submodlib library~\cite{kaushal_submodlib_2022}. More specifically the disparity min and FL acquisition functions work on the features of data obtained by computing the last hidden layer output from the model, this is known as the feature representation. For disparity min, the similarity metric used is cosine  $\text{similarity}(\vec{p}, \vec{q}) = \frac{\vec{p} \cdot \vec{q}}{\|\vec{p}\| \|\vec{q}\|}$, while for FL the distance used is also obtained via the cosine similarity $d(\vec{p}, \vec{q}) = 1 - \text{similarity}(\vec{p}, \vec{q})$. 

The experiments were conducted on a laptop equipped with a Quadro RTX 5000 16 GB GPU (with driver version 545.23.08, and CUDA version 12.3), and a 10th Gen Intel(R) Core(TM) i9-10900K CPU operating at 3.70 GHz. The system was configured with 64.0 GB of RAM and ran on the Ubuntu operating system.

\subsubsection{Hyperparameters}

We performed ablations of the query batch size $b$ and pool size $|\mathcal{D}_\text{pool}|$ hyperparameters on CIFAR10. We determined that the optimal query batch size is 800 for CIFAR10 while for the pool size, we found out that $|\mathcal{D}_\text{pool}| = 8000$ is a good balance. 
For CIFAR100, we chose $b=400$, $|\mathcal{D}_\text{pool}| = 4000$, a learning rate of 0.001, and we ran 30 acquisition rounds
For both datasets, we consider a learning rate of 0.001 and 40 epochs at each AL round.
We run $N_\text{MC-Dropout} = 5$ MC-Dropout iterations at the inference phase when required by the acquisition function. 


\subsubsection{Winning rate using $t$-test}

Assessing the superiority of different acquisition functions poses multiple challenges. Firstly, due to the intricate nature of machine learning models and diverse datasets, one has to conduct several experimental setups in order to be sure that the AL strategy considered is better generally and not only on edge cases. Secondly, due to the inherent stochastic nature of training, it is necessary to repeat the training process several times. Moreover, the iterative nature of AL involves that the acquisition function can perform well in the first rounds and worse later.
This is why we create a \textit{winning rate} which can handle the above-mentioned difficulties. To accomplish this, we evaluate strategies using pairwise comparisons, a method extensively utilized in the AL literature \cite{ash_gone_2021}, \cite{kirsch_stochastic_2023}. 
We repeat each experimental setup, with $N$ different seeds, and obtain a set of $N$ accuracy results $a_{r}=\{a_{r,1},..., a_{r,N}\}$ in each round $r$. A two-sided $t$ test is performed. Let $a_{r}^{i}$ and $a_{r}^{j}$ be the set of accuracy scores for two different AL strategies $i$ and $j$ at AL round $r$. Then, the $t$-score  is formulated as: 
\begin{equation}
\begin{gathered}
t_{r}^{ij} = \frac{\sqrt{N} \mu^{ij}_{r}}{\sigma^{ij}_{r}}\,\,\, \text{where}~ \mu^{ij}_{r} \!= \!\frac{1}{N}\sum_{l=1}^{N} \big(a_{r,l}^{i} - a_{r,l}^{j}\big)\text{,} \!\!\! \\ \text{and}\,\,\, \sigma^{ij}_{r}=\sqrt{\frac{1}{N-1}\sum_{l=1}^{N} \Big(\big(a_{r,l}^{i} - a_{r,l}^{j}\big) - \mu^{ij}_{r}\Big)}.
\label{eq:t_score}
\end{gathered}
\end{equation}
Here, the strategy $i$ is considered to beat the strategy $j$ if $t_r^{ij} > 2.776$ with $2.776$ being the critical point of $p$-value being $0.05$.
Therefore, given $R$ acquisition rounds, one can define the \textit{winning rate} formulated as follows:
\begin{equation}
{\sf win}^{ij} =  \sum_{r=1}^{R} \frac{1}{R} \mathbf{1}_{t_{r}^{ij} > \text{2.776}}
\label{eq:winning_rate}
\end{equation}
The value of the winning rate becomes 1 if the strategy $i$ beats the strategy $j$ over all AL rounds.

\subsubsection{Pairwise comparison using heatmaps}

\begin{figure}
\begin{center}
\includegraphics[width=0.35\textwidth, keepaspectratio]{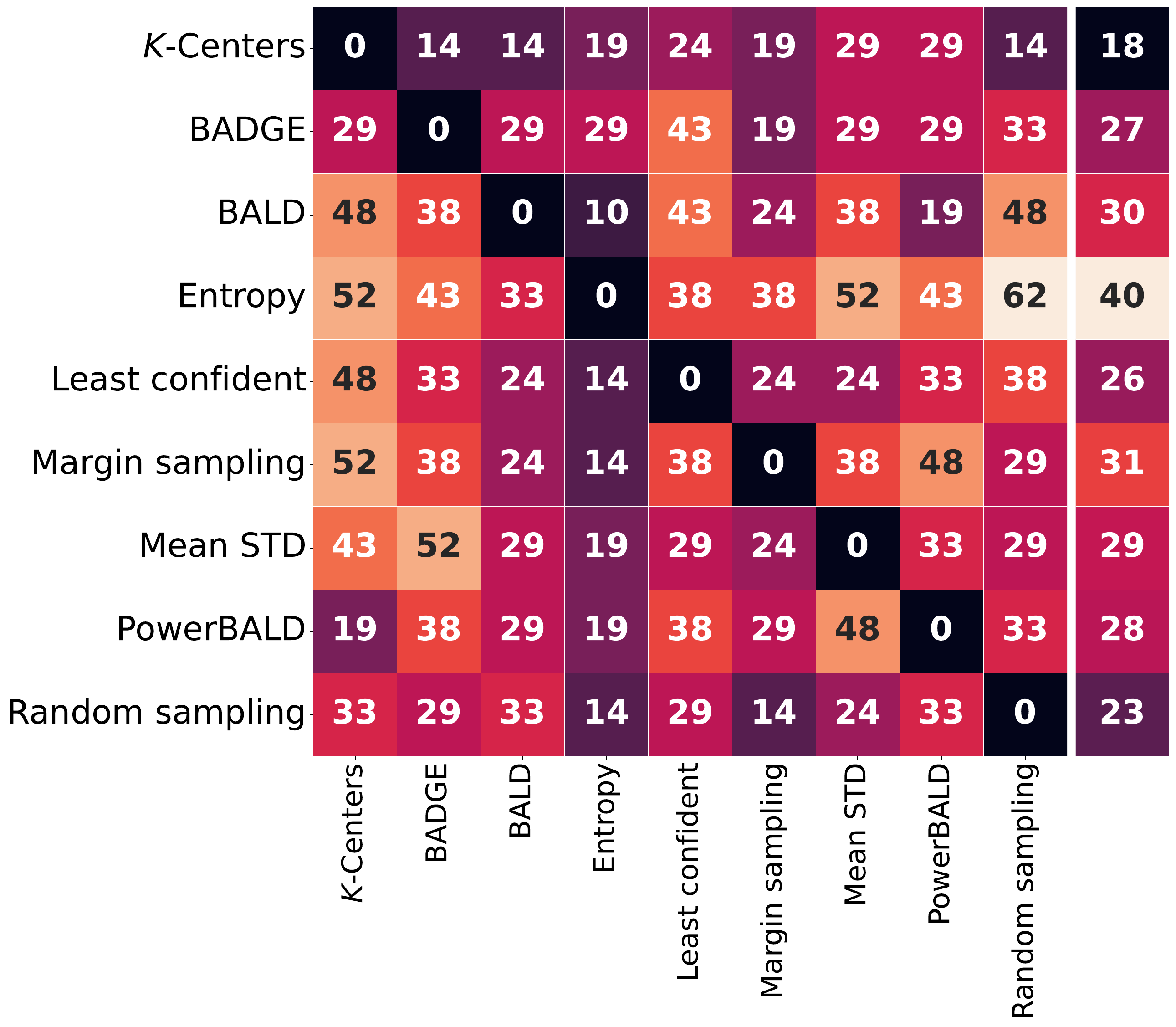}
\vspace{-3mm}
\caption{Heatmap illustrating the pairwise comparison of winning rates on CIFAR10 VGG of baseline acquisition functions. The last column represents the row average (the higher the better).}
\label{fig:CIFAR10_VGG_accuracy_comp_baseline}
\end{center}
\vspace{-5mm}
\end{figure}

In order to compare each acquisition function on each set of experiments, we plot the heatmaps (like Fig. \ref{fig:CIFAR10_VGG_accuracy_comp_baseline}) using the winning rate as introduced ahead. Therefore, one can easily compare acquisition functions. The last column on the right of the heatmap represents the mean of each row, which means that it represents on average how much the considered acquisition function wins over the others: the higher the better.

\subsection{Baseline}
\label{sec:baseline}

In order to set a baseline, we evaluate the following acquisition functions: BADGE, BALD, entropy, $K$-Centers, least confident, margin sampling, mean STD, PowerBALD, and random sampling (selecting samples randomly, like passive learning). BAIT acquisition function was not included in the baseline due to its higher computational cost and lack of accuracy gains, as our focus is on sustainable acquisition functions.

In order to highlight the relative performance, the baseline acquisition functions are compared on CIFAR10 with VGG, with the winning rates reported in Fig. \ref{fig:CIFAR10_VGG_accuracy_comp_baseline}. One can observe that $K$-Center is the worst acquisition function with an average winning rate of 18\%. Meanwhile, BADGE performs well enough in comparison to other acquisition functions with an average winning rate of 26\%. The acquisition functions based explicitly on uncertainty paradigms are hard to beat with other approaches like $K$-Centers or BADGE.

\begin{table}
\centering
\caption{Comparison of average winning rate (AWR) of baseline acquisition functions against themselves on the considered datasets with their associated rank (R).}
\label{tab:cifar_comp_dataset_model}
\small
\resizebox{0.49\textwidth}{!}{
\begin{tabular}{|@{}c|cc|cc|cc|cc@{}|}
\hline
& \multicolumn{2}{c|}{CIFAR10} & \multicolumn{2}{c|}{CIFAR10} & \multicolumn{2}{c|}{CIFAR100}  & \multicolumn{2}{c@{}|}{PTB-XL}\\
& \multicolumn{2}{c|}{VGG} & \multicolumn{2}{c|}{ResNet} & \multicolumn{2}{c|}{VGG} & \multicolumn{2}{c@{}|}{ResNet101} \\
\hline
\textbf{Acq. Func.} & \textbf{AWR} & \textbf{R} & \textbf{AWR} & \textbf{R} & \textbf{AWR} & \textbf{R} & \textbf{AWR} & \textbf{R}  \\
\hline
BADGE & 27 & 6 & 36 & 3 & 24 & 7 & 48 & 1 \\ \hline
BALD & 30 & 3 & 43 & 2 & 47 & 1 & 40 & 2 \\ \hline
Entropy & 40 & 1 & 30 & 5 & 11 & 9 & 39 & 3 \\ \hline
$K$-Centers & 18 & 9 & 9.6 & 9 & 39 & 2 & 5 & 9 \\ \hline
Least confident & 26 & 7 & 33 & 4 & 29 & 6 & 35 & 4 \\ \hline
Margin sampling & 31 & 2 & 28 & 6 & 31 & 4 & 21 & 7 \\ \hline
Mean STD & 29 & 4 & 52 & 1 & 30 & 5 & 29 & 6 \\ \hline
PowerBALD & 28 & 5 & 20 & 8 & 37 & 3 & 19 & 8 \\ \hline
Random sampling & 23 & 8 & 23 & 7 & 20 & 8 & 30 & 5 \\ \hline
\bottomrule
\end{tabular}
}
\end{table}

In Table \ref{tab:cifar_comp_dataset_model}, the comparison between the average winning rate of the baselines reveals that there is no huge rank change between the results of CIFAR10 and CIFAR100, except for $K$-Centers and entropy, when using the same model. By comparing the results on CIFAR10 using ResNet with the one using VGG, it is evident that there is a huge change in rank and average winning rate between ResNet and VGG results on CIFAR10. The PTB-XL dataset, consisting of time series data, clearly shows the ineffectiveness of $K$-Centers, while BADGE proves highly effective, reaching a 48\% average winning rate. This result highlights the ranking variations caused by domain differences with respect to image datasets. $K$-Centers relies on the assumption that feature-space distance results in informational diversity. That assumption holds better for structured visual datasets like CIFAR-10 than for noisy, high-variability biomedical signals like PTB-XL, where embeddings in early AL stages are not yet reliable for diversity-based selection.

\subsection{Results of Parallel and Hybrid structures} 
\label{sec:result_hybrid}

\begin{figure}
\begin{center}
\includegraphics[width=0.5\textwidth, keepaspectratio]{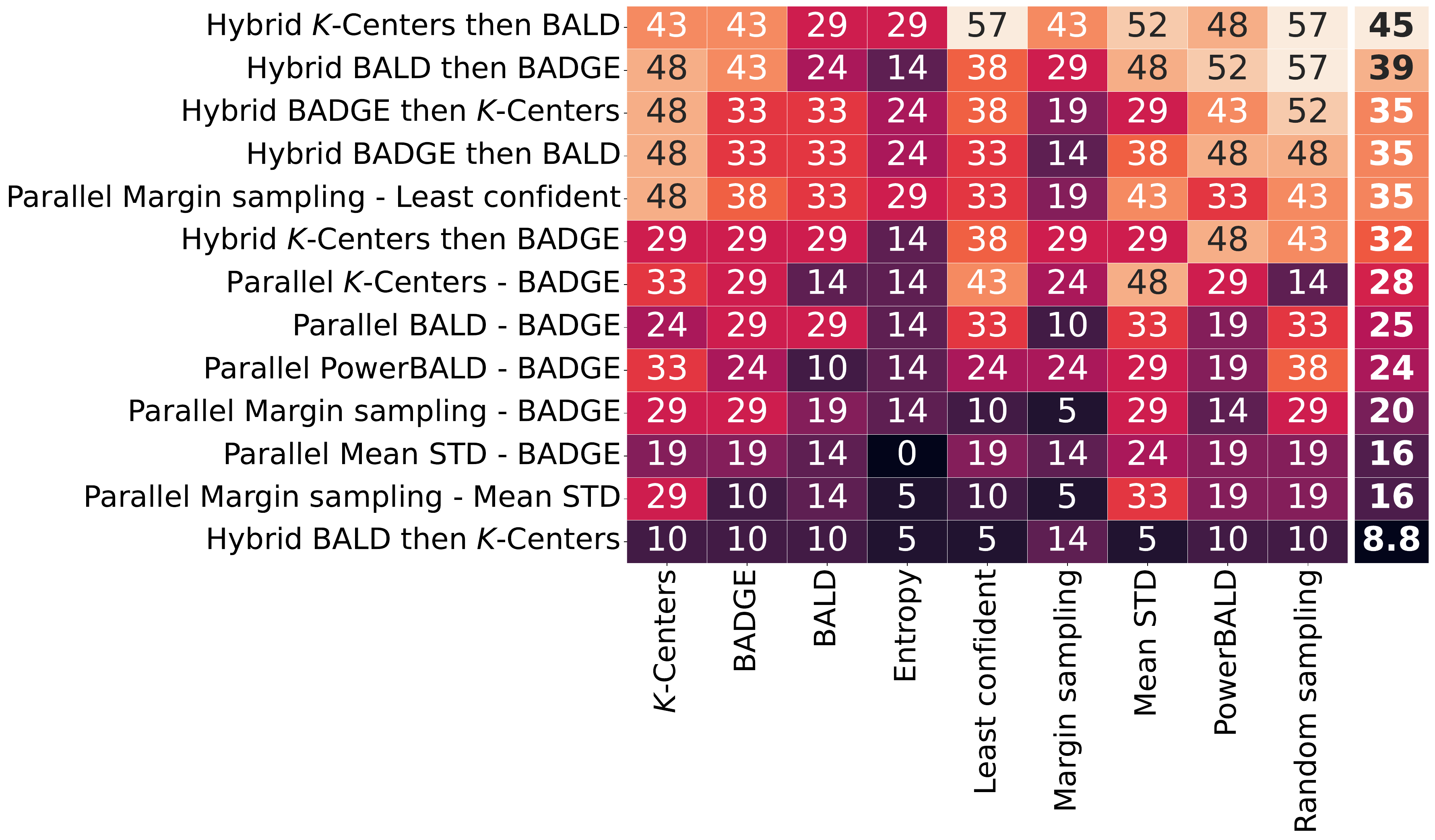}
\vspace{-5mm}
\caption{Heatmap illustrating pairwise comparison of winning rates on CIFAR10 VGG of multiple Parallel and Hybrid combinations of acquisition functions against baseline acquisition functions. 
}
\label{fig:cifar10_VGG_accuracy_comp_multiple_parallel_hybrid_comb}
\end{center}
\vspace{-5mm}
\end{figure}

For the parallel-ranked combinations, we assessed various pairs of acquisition functions on CIFAR-10 with VGG, including BALD, Margin Sampling, PowerBALD, Least Confident, and Mean STD. None of these pairs demonstrated any improvement, achieving only a 10\% average winning rate against the baseline. In fact, they significantly reduced performance compared to the original performance of each individual component in the pairs.
In the Parallel structure, we evaluated several pairs of acquisition functions on CIFAR-10 using the VGG model: BADGE and BALD, BADGE and $K$-Centers, BADGE and mean STD, BADGE and PowerBALD, BADGE and margin sampling, margin sampling and mean STD, and margin sampling and least confident.
As shown in Figure \ref{fig:cifar10_VGG_accuracy_comp_multiple_parallel_hybrid_comb}, the average winning rate of parallel combinations of acquisition functions against the baseline does not exceed 28\%. Among these combinations, \textit{Parallel $K$-Centers - BADGE} achieved the highest average performance. However, splitting the pool into two parts often leads to selecting highly similar samples, exacerbating batch selection issues.
The performance of Parallel structures does not justify the additional computational expense of running two acquisition functions.


\subsection{Results of Series structure}
\label{sec:result_series}

Here we consider the combination of two acquisition functions with the first acquisition function selecting $\kappa b = 2b$ samples on which the second acquisition function selects $b$ samples, where $b$ is the budget for one round of AL. An ablation of the $\kappa$ could be run to see if larger or smaller $\kappa$ would result in performance improvement with a time trade-off.
The series structure allows us to aggregate two acquisition functions based on two different criteria. We consider 16 different aggregations of uncertainty-based acquisition functions, namely BALD and BADGE, with representativity-based or diversity-based ones such as FL, disparity min, feature-based, $K$-Centers, or BADGE. 

\begin{figure}
\begin{center}
\includegraphics[width=0.5\textwidth, keepaspectratio]{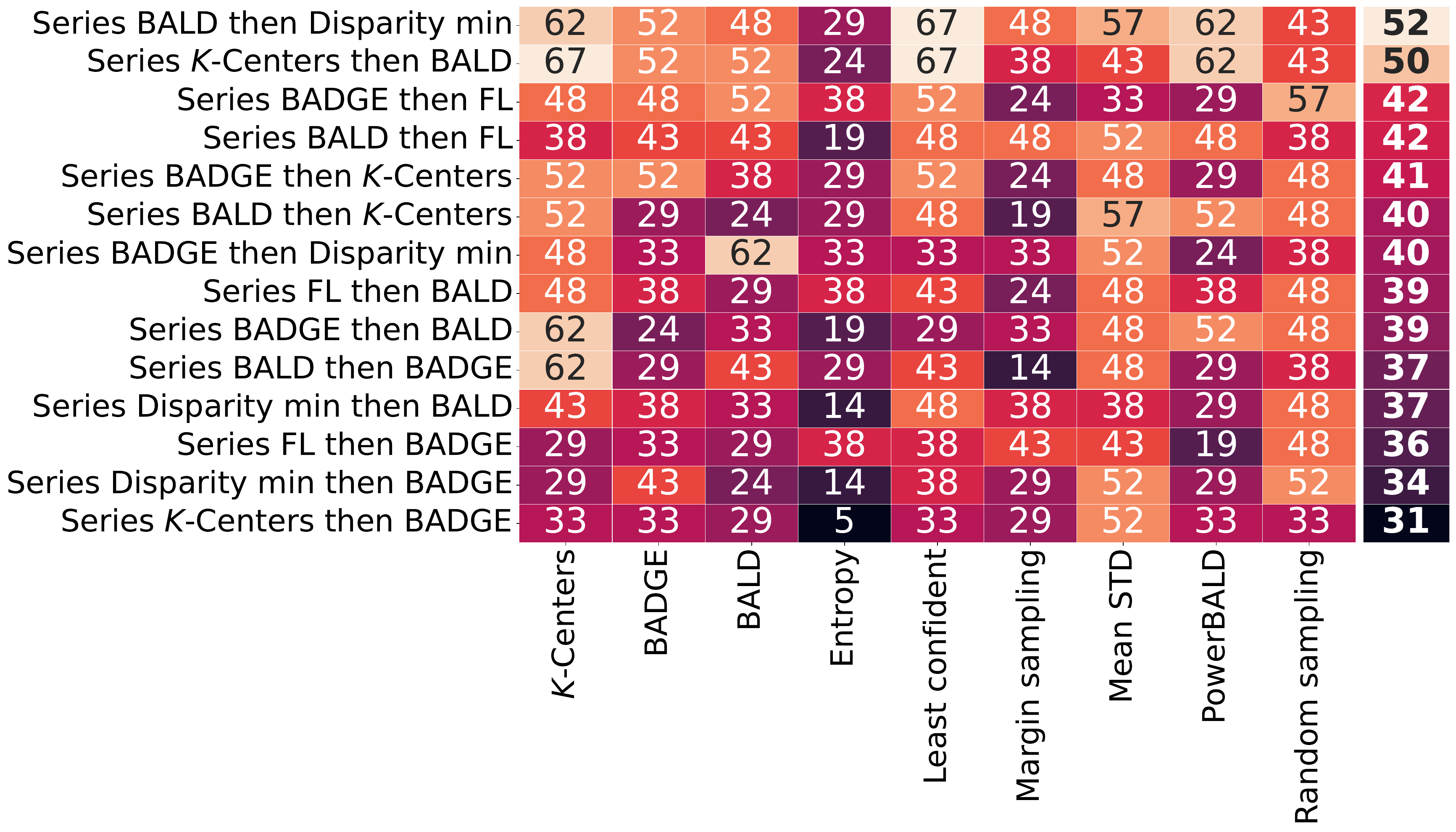}
\vspace{-6mm}
\caption{Heatmap illustrating pairwise comparison of winning rates on CIFAR10 VGG of multiple series combinations of uncertainty and representativity or diversity based acquisition functions against baseline acquisition functions. 
}
\label{fig:cifar10_VGG_accuracy_comp_multiple_series_comb}
\end{center}
\vspace{-5mm}
\end{figure}

As expected, the aggregated acquisition functions perform generally well when first running an uncertainty based function to reduce the choice of samples and then running a diversity or representativity-based function. This is shown in Fig. \ref{fig:cifar10_VGG_accuracy_comp_multiple_series_comb} for CIFAR10 VGG. The best score belongs to \textit{Series BALD then Disparity min} with an average winning rate of 52\%. This is due to the fact that BALD selects highly similar samples and uses disparity min hence allowing us to ensure diversity among the samples that are the most informative according to BALD. 

The fact that both \textit{Series BADGE then FL} and \textit{Series BALD then FL} achieve a winning rate of at least 43\% against BALD and BADGE demonstrates the effectiveness of incorporating a representativity step. The most surprising finding is that using FL after BADGE outperforms using BADGE alone, even though BADGE already includes a clustering step. This improvement may be attributed to FL leveraging the model's knowledge via the feature representation in the last hidden layer, rather than relying solely on the uncertainty derived from MC-Dropout.
Interestingly, \textit{Series BADGE then $K$-Centers} does not yield significant improvements, achieving a winning rate of only 29\% against BALD and BADGE.. 


Surprisingly, combining representativity-based acquisition functions with uncertainty-based ones also performs well. For instance, \textit{Series $K$-Centers then BALD} achieves an average winning rate of 50\%. Similarly, aggregated functions like \textit{Series BALD then Disparity min} and \textit{Series $K$-Centers then BALD} show a strong winning rate of 52\% against both BALD and BADGE. This finding is interesting as it suggests the potential to sample a larger pool and then apply algorithms to select the most representative samples.

We ran \textit{Series $K$-Centers then BALD} and \textit{Series BALD then Disparity min} which are the best Series acquisition functions on CIFAR10 using ResNet, on CIFAR100 using VGG, and on PTB-XL using ResNet101. The results are presented in Fig. \ref{fig:cifar10_resnet_accuracy_comp}, Fig. \ref{fig:cifar100_VGG_accuracy_comp} and Fig. \ref{fig:ptb-xl_accuracy_comp}. The good result of the Series structure can also be observed on CIFAR100 using VGG: \textit{Series $K$-Centers then BALD} has an average winning rate of 53\% against the baseline, while \textit{Series BALD then Disparity min} has an average winning rate of 36\% but with good performance against Entropy, BADGE or $K$-Centers. For CIFAR10 using ResNet, the results are less promising with around 30\% of average winning rate.
On the ECG classification task, which is a completely different domain from image classification, \textit{Series BALD then Disparity min} performs well, as shown in Fig. \ref{fig:ptb-xl_accuracy_comp}, with an average winning rate of 52\%. However, this is not the case for \textit{Series $K$-Centers then BALD}, which particulary underperforms with an average wining rate of 9\%. The reason is that $K$-Centers alone performs particularly poorly on the PTB-XL dataset with an average wining rate of 5\%, thereby harming the performance of the combined strategy.



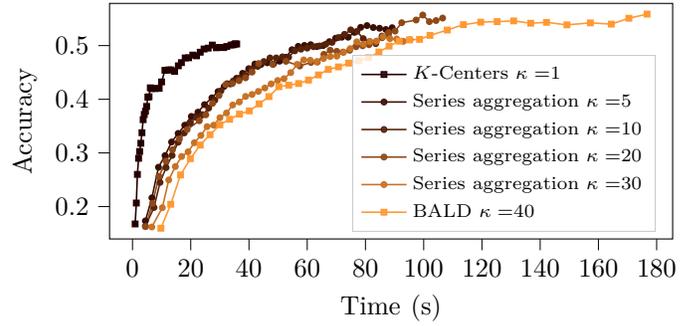
\begin{figure}[ht]
    \centering
    \input{figures/sensitivity-analysis/series} 
    \caption{Accuracy over time for Series $K$-Centers then BALD with different ratio $\kappa$ values on CIFAR10 VGG. $K$-Centers and BALD results are represented as baseline.}
    \label{fig:kappa-sensitivity}
\end{figure}

To evaluate how the $\kappa$ parameter affects the performance of the series structure, experiments were conducted using the \textit{Series $K$-Centers then BALD} approach. Twenty experiments were run for each $\kappa$ value on CIFAR-10 using VGG, with a subpool of 8,000 samples, a query batch size of 200, and 40 acquisition iterations. BALD and $K$-Centers were run as baselines for comparison. The experimental design ensures that $\kappa=1$ corresponds to running $K$-Centers alone, while $\kappa=40$ corresponds to running BALD alone. The results in Fig.~\ref{fig:kappa-sensitivity} clearly confirm our earlier findings: $K$-Centers is faster than BALD, but BALD achieves better accuracy. Consequently, \textit{Series $K$-Centers then BALD} yields better accuracy than BALD alone. The key insight from this experiment is that \textit{Series $K$-Centers then BALD} performs better with respect to both objectives (time and accuracy) with minimal sensitivity to the $\kappa$ parameter. As shown in Fig.~\ref{fig:kappa-sensitivity}, the time required to achieve the same accuracy as BALD is reduced by half in this example when choosing $\kappa=20$. It gives almost the same result as $\kappa=5$. Mostly the sub-pool size of selected by $K$-Centers has to be large enough in order for BALD to select relevant points, but large enough so that it reduces the overall computational time. 

The computational efficiency gain can be quantified by analyzing the number of model inferences required, as inference constitutes the most computationally expensive component of acquisition function evaluation. Let $N_{\text{infer}}$ denote the total number of model inferences required by an acquisition function. For BALD, $N_{\text{infer}} = N_{\text{MC-Dropout}} \cdot |\mathcal{D}_{\text{pool}}|$, while for $K$-Centers, $N_{\text{infer}} = |\mathcal{D}_{\text{pool}}| + |\mathcal{D}_{\mathcal{L}}|$. For the proposed \textit{Series $K$-Centers then BALD} approach, the total inference count is $N_{\text{infer}} = |\mathcal{D}_{\text{pool}}| + |\mathcal{D}_{\mathcal{L}}| + \kappa b  N_{\text{MC-Dropout}}$, where $\kappa b \ll |\mathcal{D}_{\text{pool}}|$. Beyond the computational speedup, $K$-Centers is specifically designed to exploit sample representativity, thereby promoting sample diversity. This enhanced diversity may account for the observed accuracy improvements relative to BALD alone.

\subsection{Results of Hybrid structure} 

In order to evaluate the effectiveness of the hybrid structure, we combine BALD, BADGE, and $K$-Centers with each other, resulting in 6 acquisition functions to discuss. 
From the first glance at Fig. \ref{fig:cifar10_VGG_accuracy_comp_multiple_parallel_hybrid_comb}, one can observe that the hybrid combinations are better than the parallel ones with higher winning rates against the baselines. However, it is apparent that this type of structure does not compare favorably to the series structure. For instance, \textit{Hybrid BADGE then $K$-Centers} achieves an average winning rate of 35\% against the baseline, whereas \textit{Series BADGE then $K$-Centers} achieves 40\%. Moreover, the Hybrid structure incurs higher costs than the Series structure. In the Series structure, the second acquisition function operates on samples selected within the batch by the first acquisition function, whereas in the Hybrid structure, the second function operates on the entire pool excluding the previously selected samples, which is more costly.
Although the Hybrid structure mitigates the disadvantage of the Parallel structure, which splits the pool into two parts resulting in the selection of highly similar samples, the results do not exceed those of series combinations. Consequently, the Hybrid structure does not justify its cost compared to the Series combination.

\subsection{Results of Adaptive feedback structure and of structures switching between acquisition functions}
\label{sec:result_alternating}

\begin{figure}
\begin{center}
\includegraphics[width=0.5\textwidth, keepaspectratio]{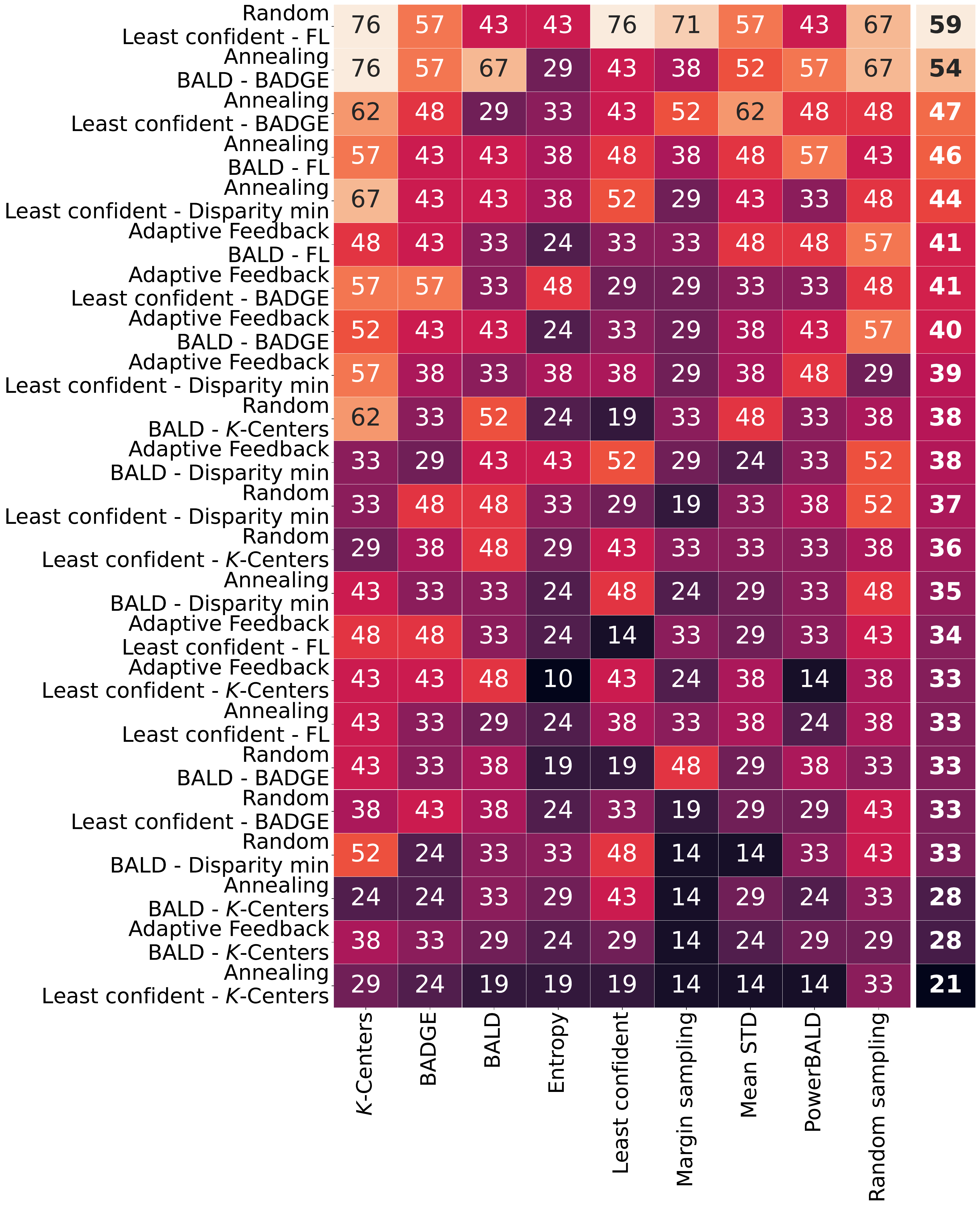}
\vspace{-8mm}
\caption{Heatmap illustrating pairwise comparison of winning rates on CIFAR10 VGG of multiple Adaptive feedback combinations against baseline acquisition functions. 
}
\label{fig:cifar10_VGG_accuracy_comp_adaptive_comb}
\end{center}
\vspace{-5mm}
\end{figure}

To evaluate the combination of acquisition functions, we selected three uncertainty-based functions: BALD, entropy, and least confident. These were combined with two diversity-based functions: BADGE and disparity min, as well as two representativeness-based functions: $K$-Centers and FL. For each dataset, we evaluate the performance of the following acquisition functions: random sampling, BALD, PowerBALD, $K$ Centers, BADGE, mean STD, max entropy, least confident. We benchmark combinations using adaptive feedback with loss of the previous model on the selected batch, random and annealing structures. Please note that for the annealing structure, we have chosen $T_{\text{initial exploration}} = 5$, $r = 1.5$, ${T_{\text{exploit}}}_1 = 5$, ${T_{\text{explore}}} = 5$, as we run VGG on CIFAR10 for 20 rounds only here, this choice makes it such that exploration strategy is chosen for the 5 first rounds, then exploitation for the 5 next, then exploration for the 5 next, and finally exploitation for the 5 next. On CIFAR10 with VGG, we evaluated the combinations against the baseline in Fig. \ref{fig:cifar10_VGG_accuracy_comp_adaptive_comb}.
In Fig. \ref{fig:cifar10_VGG_accuracy_comp_adaptive_comb}, it is evident that the three top combinations are \textit{Random Least confident - FL}, \textit{Annealing BALD - BADGE} and \textit{Annealing Least Confident - BADGE}. These three combinations outperform on average with more than 54\% winning rate over the baseline.

Here we considered BADGE as the exploration function (as it is based on the diversity of gradients) and BALD as the exploitation function (as it is an uncertainty-based method). However, the acquisition functions  \textit{Random BALD - BADGE} and in particular \textit{Annealing BALD - BADGE} works better than \textit{Adaptive feedback BALD - BADGE} with 50\% average winning rate on CIFAR10 for VGG. This suggests that simply alternating between BADGE and BALD with annealing or random structure could be more important than alternating between BADGE and BALD at the right moment with adaptive feedback.

This result remains surprising, considering that both BADGE and BALD aim to maximize the same objective — expected information gain, as outlined in \cite{kirsch_unifying_2022}. The main distinction lies in BADGE's design, which addresses BMAL pathologies by emphasizing sample representativeness through $k$-Means$++$ clustering on gradient embeddings, categorizing it as a diversity-based acquisition function. This emphasis on clustering might explain why combinations of BADGE and BALD achieve a winning rate of more than 50\% over BALD, which operates in a Top-$K$ manner. However, it does not fully elucidate why it outperforms BADGE alone.
One plausible interpretation is rooted in the differing operational spaces: BALD operates in prediction space (via MC-Dropout), while BADGE operates in weight space (via gradient embedding). Alternating between these spaces may synergistically enhance the final acquisition function's efficacy. This approach effectively addresses the problem from two distinct perspectives, potentially leading to improved overall performance.

\begin{figure}
\begin{center}
\includegraphics[width=0.5\textwidth, keepaspectratio]{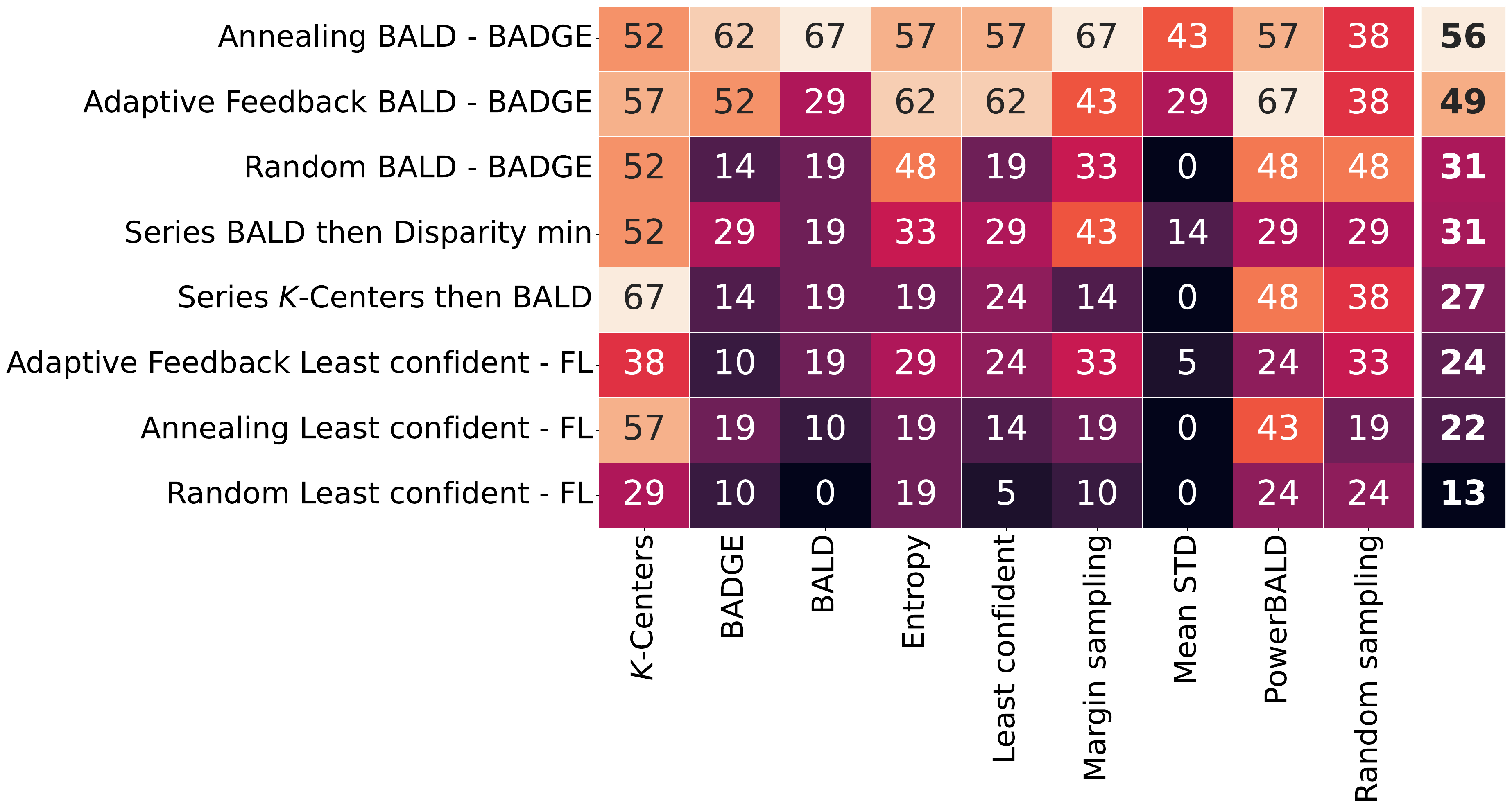}
\vspace{-8mm}
\caption{Heatmap illustrating pairwise comparison of winning rates on CIFAR10 using ResNet of multiple combinations of acquisition function against baseline acquisition functions. 
}
\label{fig:cifar10_resnet_accuracy_comp}
\end{center}
\end{figure}
\begin{figure}
\begin{center}
\includegraphics[width=0.5\textwidth, keepaspectratio]{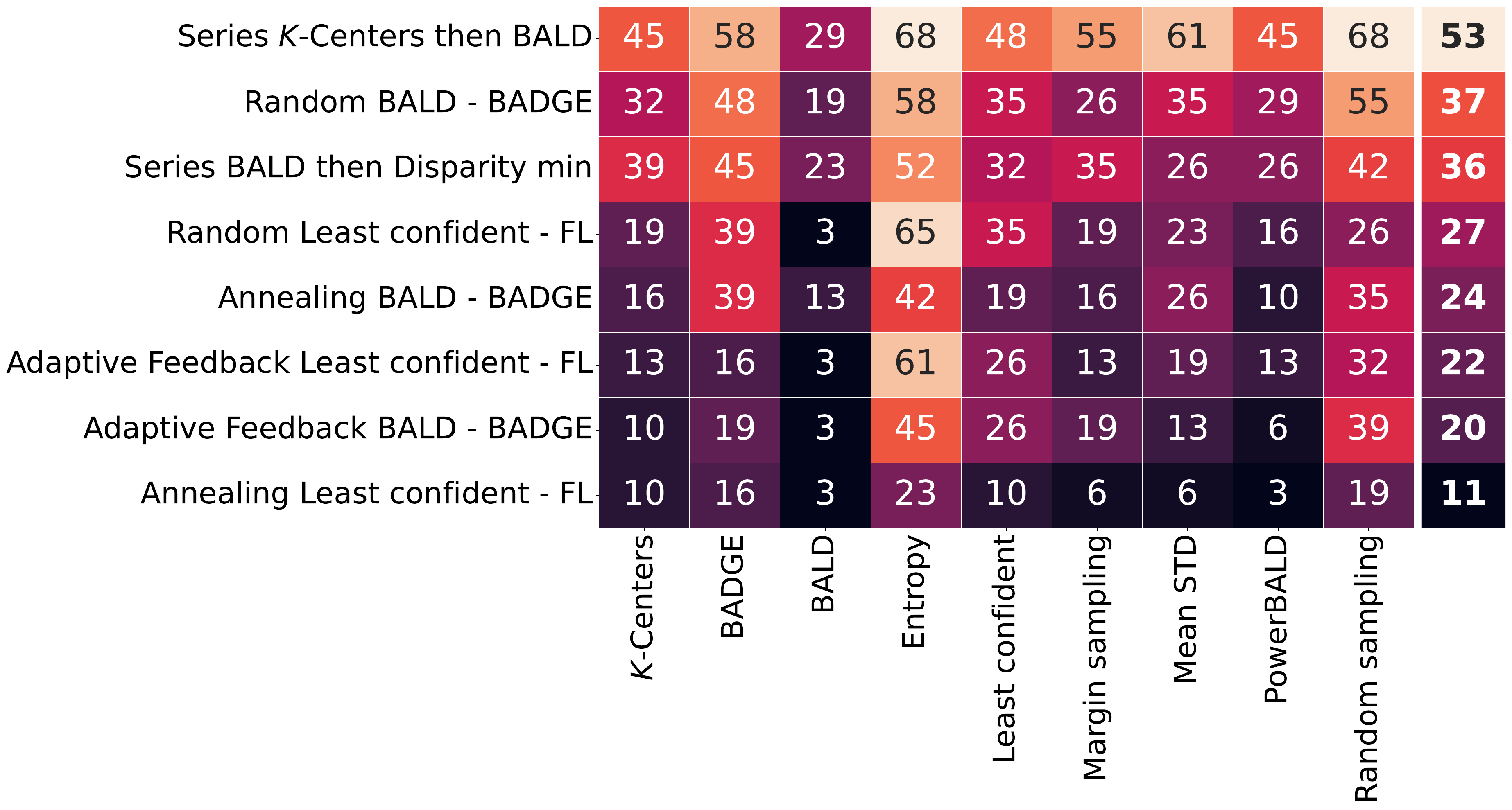}
\vspace{-8mm}
\caption{Heatmap illustrating pairwise comparison of winning rates on CIFAR100 using VGG of multiple combinations of acquisition function against baseline acquisition functions. 
}
\label{fig:cifar100_VGG_accuracy_comp}
\end{center}
\vspace{-5mm}
\end{figure}
\begin{figure}
\begin{center}
\includegraphics[width=0.5\textwidth, keepaspectratio]{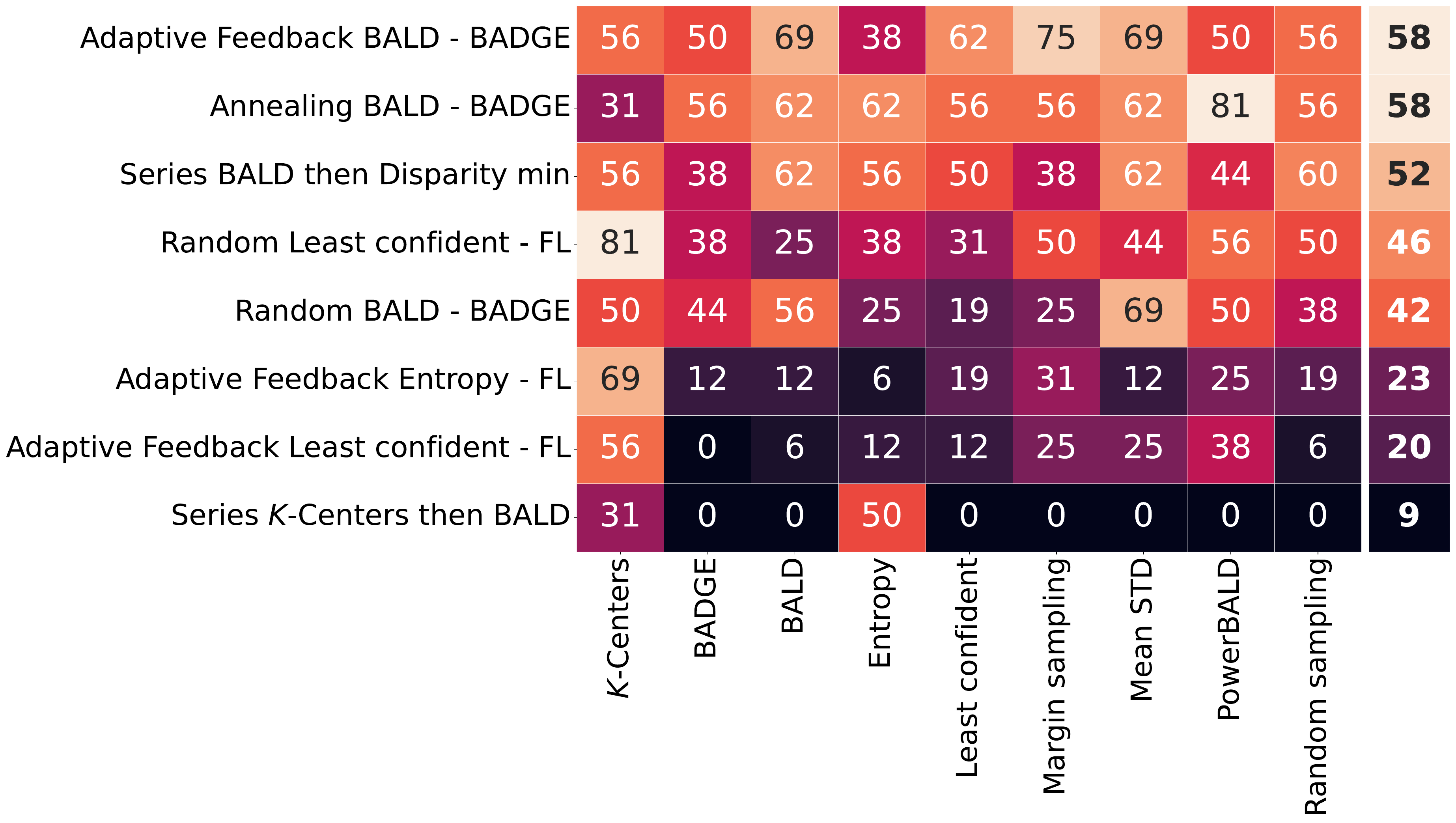}
\vspace{-8mm}
\caption{Heatmap illustrating pairwise comparison of winning rates on PTB-XL using ResNet101 of multiple combinations of acquisition function against baseline acquisition functions.
}
\label{fig:ptb-xl_accuracy_comp}
\end{center}
\vspace{-5mm}
\end{figure}

To observe if the results are stable while changing the dataset and model, we run \textit{Adaptive Feedback BALD - BADGE}, \textit{Annealing BALD - BADGE}, \textit{Random BALD - BADGE}, \textit{Adaptive Feedback Least confident - FL}, \textit{Annealing Least confident - FL}, and \textit{Random Least confident - FL} on CIFAR10 using ResNet on CIFAR100 using VGG, and on PTB-XL using ResNet101. The results are presented in Fig. \ref{fig:cifar10_resnet_accuracy_comp}, in Fig. \ref{fig:cifar100_VGG_accuracy_comp}, and in Fig. \ref{fig:ptb-xl_accuracy_comp}. The good result of the alternating structure can also be observed on CIFAR10 using ResNet: with \textit{Adaptive Feedback BALD - BADGE} and \textit{Annealing BALD - BADGE} achieving 56\% and 49\% of average winning rate respectively.
For CIFAR100 using VGG, the results are less promising with the best alternating acquisition function achieving an average winning rate of 37\% but with good performance against Entropy, and BADGE.
Although ECG classification represents a completely different domain from image classification, Fig. \ref{fig:ptb-xl_accuracy_comp} shows that \textit{Adaptive Feedback BALD - BADGE} and \textit{Annealing BALD – BADGE} are again the top acquisition functions, each attaining an average winning rate of 58\%.

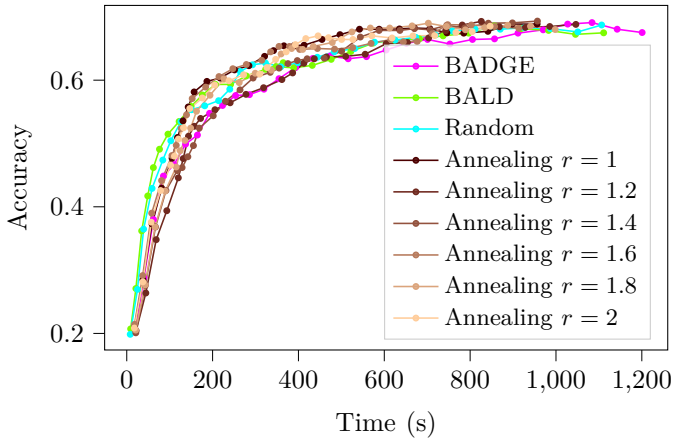
\begin{figure}[ht]
    \centering
    \input{figures/sensitivity-analysis/annealing} 
    \caption{Accuracy over time for Annealing BALD BADGE with different annealing rate $r$ values on CIFAR10 VGG. Random BALD BADGE, BALD and BADGE results are represented as baseline.}
    \label{fig:periodrate-sensitivity}
\end{figure}

To investigate the influence of the period rate $r$ on the Annealing structure, five values of $r$ ranging from $1$ to $2$ were evaluated. For each value, twenty experiments were conducted on CIFAR-10 using VGG, with a subpool size of $5{,}000$ samples, a query batch size of $500$, and $40$ acquisition iterations. The results are shown in Fig.~\ref{fig:periodrate-sensitivity}. The settings $r=1$ and $r=1.2$ yielded the highest performance, although the results for $r=2$ were comparable in terms of both accuracy and runtime. Notably, $r=1$ corresponds here to the case where the exploration and exploitation acquisition functions operates equally, the result should be similar to \textit{Random BALD - BADGE}. However, starting with BADGE for a predefined number of iterations $T_{\text{initial exploration}}$, and restarting for $T_{\text{exploration}}$ iterations after $T_{\text{exploitation}}$ iterations appears to produce significant improvements.


\subsection{Robustness and energy gain of aggregated acquisition functions}

\begin{table*}
\caption{Comparison of acquisition functions when first having reached 80\% of accuracy for CIFAR10 using VGG averaged in 12 repetitions of AL process (AL processes are stopped after reaching 80\% of accuracy or after one hour of execution)}
\resizebox{\textwidth}{!}{
\centering
\begin{tabular}{|l|l|l|l|l|l|l|}
\hline
Acquisition function used & Execution duration & Acq. round & \# samples & \% of dataset & Final accuracy & \# goal reached \\
\hline
Entropy & 38min 7s & 30 & 24000 & 40\% & 80\% & 3 out of 12\\
BALD & 55min 26s & 35 & 28000 & 46\% & 81\% & 2 out of 12\\
Least confident & 35min 5s & 28 & 22666 & 37\% & \textbf{82\%} & 3 out of 12\\
PowerBALD & 35min 58s & 27 & 22133 & 36\% & 80\% & 3 out of 12\\
$K$-Centers & \textbf{28min 21s} & 26 & 20800 & 34\% & 80\% & 2 out of 12\\
BADGE & 47min 12s & 31 & 24960 & 41\% & 81\% & 5 out of 12\\
Margin sampling & 38min 44s & 29 & 23800 & 39\% & 81\% & 4 out of 12\\
\textit{Series $K$-Centers then BALD} & 34min 21s & \textbf{26} & \textbf{21333} & \textbf{35\%} & 80\% & \textbf{6 out of 12}\\
\textit{Adaptive feedback BALD - BADGE} & 39min 12s & 31 & 25066 & 41\% & 80\% & 3 out of 12\\
\textit{Random BALD - BADGE} & 40min 51s & 29 & 23866 & 39\% & 80\% & \textbf{6 out of 12}\\
\textit{Annealing BALD - BADGE} & 45min 34s & 32 & 25733 & 42\% & 80\% & \textbf{6 out of 12}\\
\hline
\end{tabular}
}
\label{tab:robustness_averaged}
\end{table*}

In order to assess the execution duration for reaching a particular value of accuracy, we run the same experiment (CIFAR10 and VGG, 40 epochs of training at each round, a query batch size of 800, a pool size of 8000) until the model reaches 80\% of accuracy or until the AL process excess one hour of execution time. The averaged results of 12 repetitions of experiments are compiled in \Cref{tab:robustness_averaged}. It reveals that in terms of execution duration, $K$-Centers finishes on average the first with 28 minutes the first when it is able to reach 80\% within one hour. However, $K$-Centers finishes before one hour only 2 times over 12 repetitions of AL process. Executing BALD on the subpool extracted using $K$-Centers brings this number from 2 successes to 6 successes over 12 repetitions. When \textit{Series $K$-Centers then BALD} succeeds reaching 80\% of accuracy, able to do it in 9 acquisition rounds less than with BALD alone. In terms of labeling cost, this represents $9 \times 800 = 7200$ labels less, as a query batch size of $b = 800$ was used. Regarding the entirety of CIFAR10, this represents 12\% of the total available samples. This means that for the same accuracy goal, one needs 12\% of energy less to train the model. Actually, the cost of energy for training a model depends directly on the number of samples selected. 

The methods combining BALD and BADGE are also more robust in terms of reaching 80\% accuracy before one hour, achieving success 6 times over 10 for \textit{Random BALD - BADGE} and \textit{Annealing BALD - BADGE}. They are also better in terms of execution time than BALD and BADGE alone. Regarding the energy cost, the same argument can be developed for the aggregated functions \textit{Random BALD - BADGE} and \textit{Annealing BALD - BADGE}. However, our experiment did not reveal the same improvement range in number of rounds to reach the same objective as from using \textit{Series $K$-Centers then BALD} instead of $K$-Centers or BALD alone, but it still improves the robustness of the acquisition functions. 

On top of this data-efficiency gain, we already discussed the potential gain in terms of the execution duration of the series and adaptive feedback structure. In fact, we have shown that the series structure if the acquisition functions are well chosen can lead to a reduction in execution time compared to one alone without loss of accuracy. This is the case for \textit{Series $K$-Centers then BALD} which requires fewer acquisition rounds. The difference arises from the fact that BALD requires $N_\text{MC dropout}|\mathcal{D}_\text{pool}|$ forward passes, whereas $K$-Centers requires only $|\mathcal{D}_\text{pool}|$ forward passes. Applying BALD on top of $K$-Centers requires only $|\mathcal{D}_\text{pool}|+N_\text{MC dropout} b << N_\text{MC dropout}|\mathcal{D}_\text{pool}|$ forward passes.
 Moreover, concerning adaptive feedback structure, there is at least no escalation of the running time of the acquisition functions itself. Our proposed structures, therefore, at least, do not lengthen the execution time of acquisition functions, and in certain scenarios, even shorten it. Considering that energy consumption is proportional to the execution time, our proposed structures also do not escalate the energy costs of active learning processes, and in some cases, they even reduce it. We experimentally demonstrated that our proposed structures can achieve data-efficiency gain without increasing the overhead computation cost of AL.

By wisely selecting the structures and the acquisition functions, we are able to obtain gain in terms of both the quality of the data selected and the efficiency of the method to select them, leading to an overall energy gain.

\subsection{Key practical insights}
Based on the comprehensive experimental results presented in this work across various datasets and model architectures, several key factors emerge as critical determinants of AL performance and the effectiveness of aggregated acquisition functions in terms of accuracy:
\begin{itemize}
    \item \textbf{Model architecture} interactions play a crucial role, as evidenced by the substantial changes in ranking between VGG and ResNet results on CIFAR-10, as discussed in Section \ref{sec:baseline}.
    \item \textbf{Batch mode pathologies} significantly affect performance: Top-$K$ selection tends to amplify redundancy issues in parallel-ranked structures (see Section \ref{sec:result_hybrid}), whereas series structures mitigate these problems through progressive filtering (see Section \ref{sec:result_series}). 
    \item \textbf{Alternating strategies}, such as switching between multiple methods, yet simple prove to be essential: approaches alternating between BADGE and BALD consistently achieve winning rates above 50\% across domains, as they operate in complementary spaces (weight vs. prediction). This is observed for \textit{Random BALD–BADGE}, \textit{Annealing BALD–BADGE}, and \textit{Adaptive Feedback BALD–BADGE} (see Section \ref{sec:result_alternating}). In a similar vein, the results of \textit{Hybrid BADGE then BALD} presented in Section \ref{sec:result_hybrid} can also be interpreted as stemming from this complementary behavior.
    \item \textbf{Domain} appears to be a key determinant for some acquisition functions: the feature space assumptions underlying diversity-based methods like $K$-Centers may not translate effectively to time series data, where early-stage embeddings are often unreliable, as shown in Section \ref{sec:baseline}. Therefore the superior performance of \textit{Series $K$-Centers then BALD} on structured visual datasets such as CIFAR-10/100 (achieving winning rates above 50\%) contrasts sharply with its poor performance on the noisy, high-variability PTB-XL ECG dataset (9\% winning rate) as denoted in Section \ref{sec:result_series}.
\end{itemize}
These findings indicate that successful aggregation depends on careful consideration of dataset characteristics, architectural compatibility, and the underlying principles of the constituent acquisition functions, rather than on a one-size-fits-all combination strategy.
While most findings were validated across CIFAR10, CIFAR100, and PTB-XL, the relative strength of each aggregation structure varied. This reinforces the need to test acquisition functions in the target domain rather than relying on benchmarks alone.

To improve performance in terms of computation time while keeping the number of acquisition iterations constant, only the series structure was able to reduce computational time by progressively decreasing the pool size (see Section \ref{sec:result_series}).

\section{Conclusion and Future Works}

Active learning is a powerful technique that can reduce the number of samples necessary for model training to achieve a specified accuracy. This reduction not only lowers labeling costs but also leads to energy savings during the training of neural network models. Therefore, it is a learning method that can promote both data efficiency and energy efficiency.
In this paper, we implemented and compared various state-of-the-art acquisition functions in terms of accuracy and computational costs. 
To address the exploration-exploitation dilemma, we introduced six types of aggregation structures: series, parallel, hybrid, adaptive feedback, random exploration, and annealing exploration. 
After conducting ablation studies on hyperparameters, we compared the aggregated acquisition functions to a set of state-of-the-art baseline acquisition functions in terms of both energy cost and accuracy.


In these comparisons, two predominant structural types emerged. The series structure, in particular, showed a reduction in computational cost by using one acquisition function to initially reduce the pool size before applying the second one. Notably, combinations such as $K$-Centers followed by BALD demonstrated superior performance compared to BALD alone in two dataset-model pairs. This approach reduced the acquisition cost by almost half and decreased the number of labels needed to reach 80\% accuracy by 12\%.
On the other hand, structures that alternate between exploration-based and exploitation-based acquisition functions at each round—whether based on a metric (adaptive feedback), the round number (annealing), or randomly (random)—have shown robust results. 


In future work, we aim to address the query batch size parameter, $b$. Implementing a flexible query batch size could lead to further improvements in energy efficiency. This might involve adjusting the query batch size through an annealing process, initially exploring more to refine the model and accurately capture boundary decisions using acquisition functions. Another approach could involve treating $b$ as a penalty term in an optimization process or using the information score of submodular functions to measure the benefit of adding a sample, ultimately excluding it if it is not valuable enough. 



\bibliographystyle{IEEEtran}
\bibliography{collection}

\end{document}

%% file: figures/structure/figure_parallel_ranking_small.tex
\tikzset{every picture/.style={line width=0.75pt}} 

\begin{tikzpicture}[x=0.75pt,y=0.75pt,yscale=-1,xscale=1]

\draw    (112,147) -- (112,177) ;
\draw [shift={(112,179)}, rotate = 270] [color={rgb, 255:red, 0; green, 0; blue, 0 }  ][line width=0.75]    (10.93,-3.29) .. controls (6.95,-1.4) and (3.31,-0.3) .. (0,0) .. controls (3.31,0.3) and (6.95,1.4) .. (10.93,3.29)   ;
\draw   (8,115) -- (216,115) -- (216,147) -- (8,147) -- cycle ;
\draw    (111,24) -- (111,48) ;
\draw    (31,56) -- (111,48) ;
\draw    (111,48) -- (183,56) ;
\draw   (128,80) -- (216,80) -- (216,99) -- (128,99) -- cycle ;
\draw    (183,56) -- (183,78) ;
\draw [shift={(183,80)}, rotate = 270] [color={rgb, 255:red, 0; green, 0; blue, 0 }  ][line width=0.75]    (10.93,-3.29) .. controls (6.95,-1.4) and (3.31,-0.3) .. (0,0) .. controls (3.31,0.3) and (6.95,1.4) .. (10.93,3.29)   ;
\draw    (31,56) -- (31,78) ;
\draw [shift={(31,80)}, rotate = 270] [color={rgb, 255:red, 0; green, 0; blue, 0 }  ][line width=0.75]    (10.93,-3.29) .. controls (6.95,-1.4) and (3.31,-0.3) .. (0,0) .. controls (3.31,0.3) and (6.95,1.4) .. (10.93,3.29)   ;
\draw   (10,80) -- (98,80) -- (98,99) -- (10,99) -- cycle ;
\draw    (184,99) -- (184,113) ;
\draw [shift={(184,115)}, rotate = 270] [color={rgb, 255:red, 0; green, 0; blue, 0 }  ][line width=0.75]    (10.93,-3.29) .. controls (6.95,-1.4) and (3.31,-0.3) .. (0,0) .. controls (3.31,0.3) and (6.95,1.4) .. (10.93,3.29)   ;
\draw    (32,99) -- (32,113) ;
\draw [shift={(32,115)}, rotate = 270] [color={rgb, 255:red, 0; green, 0; blue, 0 }  ][line width=0.75]    (10.93,-3.29) .. controls (6.95,-1.4) and (3.31,-0.3) .. (0,0) .. controls (3.31,0.3) and (6.95,1.4) .. (10.93,3.29)   ;

\draw (41,117) node [anchor=north west][inner sep=0.75pt]  [font=\footnotesize] [align=left] {\begin{minipage}[lt]{102.51pt}\setlength\topsep{0pt}
\begin{center}
Sum rank of both \\acquisition functions scores
\end{center}

\end{minipage}};
\draw (117,153) node [anchor=north west][inner sep=0.75pt]  [font=\small] [align=left] {\textit{argmax}};
\draw (102,184) node [anchor=north west][inner sep=0.75pt]  [font=\footnotesize] [align=left] {$\mathcal{D}_t^*$};
\draw (97,8) node [anchor=north west][inner sep=0.75pt]  [font=\footnotesize] [align=left] {$\mathcal{D}_{\text{pool}}$};
\draw (130.67,85) node [anchor=north west][inner sep=0.75pt]  [font=\footnotesize] [align=left] {Acq. function 2};
\draw (116,29) node [anchor=north west][inner sep=0.75pt]  [font=\footnotesize] [align=left] {\textit{Share same pool}};
\draw (13,85) node [anchor=north west][inner sep=0.75pt]  [font=\footnotesize] [align=left] {Acq. function 1};

\end{tikzpicture}

%% file: figures/structure/figure_parallel_small.tex
\tikzset{every picture/.style={line width=0.75pt}} 

\begin{tikzpicture}[x=0.75pt,y=0.75pt,yscale=-1,xscale=1]

\draw    (128,24) -- (128,48) ;
\draw    (48,56) -- (128,48) ;
\draw    (128,48) -- (200,56) ;
\draw   (145,80) -- (233,80) -- (233,104) -- (145,104) -- cycle ;
\draw    (200,56) -- (200,78) ;
\draw [shift={(200,80)}, rotate = 270] [color={rgb, 255:red, 0; green, 0; blue, 0 }  ][line width=0.75]    (10.93,-3.29) .. controls (6.95,-1.4) and (3.31,-0.3) .. (0,0) .. controls (3.31,0.3) and (6.95,1.4) .. (10.93,3.29)   ;
\draw    (48,104) -- (48,120) ;
\draw    (200,104) -- (200,120) ;
\draw    (128,120) -- (200,120) ;
\draw    (48,120) -- (128,120) ;
\draw    (128,120) -- (128,142) ;
\draw [shift={(128,144)}, rotate = 270] [color={rgb, 255:red, 0; green, 0; blue, 0 }  ][line width=0.75]    (10.93,-3.29) .. controls (6.95,-1.4) and (3.31,-0.3) .. (0,0) .. controls (3.31,0.3) and (6.95,1.4) .. (10.93,3.29)   ;
\draw    (48,56) -- (48,78) ;
\draw [shift={(48,80)}, rotate = 270] [color={rgb, 255:red, 0; green, 0; blue, 0 }  ][line width=0.75]    (10.93,-3.29) .. controls (6.95,-1.4) and (3.31,-0.3) .. (0,0) .. controls (3.31,0.3) and (6.95,1.4) .. (10.93,3.29)   ;
\draw   (27,80) -- (115,80) -- (115,104) -- (27,104) -- cycle ;

\draw (113,10) node [anchor=north west][inner sep=0.75pt]  [font=\footnotesize] [align=left] {$\mathcal{D}_{\text{pool}}$};
\draw (147.67,85) node [anchor=north west][inner sep=0.75pt]  [font=\footnotesize] [align=left] {Acq. function 2};
\draw (105,106) node [anchor=north west][inner sep=0.75pt]  [font=\small] [align=left] {\textit{Concat}};
\draw (130,27) node [anchor=north west][inner sep=0.75pt]  [font=\footnotesize] [align=left] {\textit{Split pool}};
\draw (119,149) node [anchor=north west][inner sep=0.75pt]  [font=\footnotesize] [align=left] {$\mathcal{D}_t^*$};
\draw (30,85) node [anchor=north west][inner sep=0.75pt]  [font=\footnotesize] [align=left] {Acq. function 1};

\end{tikzpicture}

%% file: figures/structure/figure_annealing_small.tex
\tikzset{every picture/.style={line width=0.75pt}} 

\begin{tikzpicture}[x=0.75pt,y=0.75pt,yscale=-1,xscale=1]

\draw    (179.46,19.91) -- (179.46,40.2) ;
\draw    (124.58,55.42) -- (179.46,40.2) ;
\draw    (124.58,55.42) -- (124.36,85.55) ;
\draw [shift={(124.35,87.55)}, rotate = 270.41] [color={rgb, 255:red, 0; green, 0; blue, 0 }  ][line width=0.75]    (10.93,-3.29) .. controls (6.95,-1.4) and (3.31,-0.3) .. (0,0) .. controls (3.31,0.3) and (6.95,1.4) .. (10.93,3.29)   ;
\draw   (190.48,87.55) -- (272,87.55) -- (272,121.37) -- (190.48,121.37) -- cycle ;
\draw    (229.06,53.73) -- (229.06,85.55) ;
\draw [shift={(229.06,87.55)}, rotate = 270] [color={rgb, 255:red, 0; green, 0; blue, 0 }  ][line width=0.75]    (10.93,-3.29) .. controls (6.95,-1.4) and (3.31,-0.3) .. (0,0) .. controls (3.31,0.3) and (6.95,1.4) .. (10.93,3.29)   ;
\draw   (86.92,87.55) -- (168.44,87.55) -- (168.44,121.37) -- (86.92,121.37) -- cycle ;
\draw    (123.2,121.37) -- (123,139) ;
\draw    (232,123) -- (232,139) ;
\draw    (178.11,139) -- (232,139) ;
\draw    (123,139) -- (178.11,139) ;
\draw    (184,139) -- (184,153) ;
\draw [shift={(184,155)}, rotate = 270] [color={rgb, 255:red, 0; green, 0; blue, 0 }  ][line width=0.75]    (10.93,-3.29) .. controls (6.95,-1.4) and (3.31,-0.3) .. (0,0) .. controls (3.31,0.3) and (6.95,1.4) .. (10.93,3.29)   ;
\draw  [fill={rgb, 255:red, 0; green, 0; blue, 0 }  ,fill opacity=1 ] (121.82,55.42) .. controls (121.82,53.55) and (123.06,52.04) .. (124.58,52.04) .. controls (126.1,52.04) and (127.33,53.55) .. (127.33,55.42) .. controls (127.33,57.29) and (126.1,58.8) .. (124.58,58.8) .. controls (123.06,58.8) and (121.82,57.29) .. (121.82,55.42) -- cycle ;
\draw  [fill={rgb, 255:red, 0; green, 0; blue, 0 }  ,fill opacity=1 ] (176.24,40.2) .. controls (176.24,38.33) and (177.48,36.82) .. (179,36.82) .. controls (180.52,36.82) and (181.76,38.33) .. (181.76,40.2) .. controls (181.76,42.07) and (180.52,43.58) .. (179,43.58) .. controls (177.48,43.58) and (176.24,42.07) .. (176.24,40.2) -- cycle ;
\draw  [fill={rgb, 255:red, 0; green, 0; blue, 0 }  ,fill opacity=1 ] (226.53,53.73) .. controls (226.53,51.86) and (227.77,50.35) .. (229.29,50.35) .. controls (230.81,50.35) and (232.04,51.86) .. (232.04,53.73) .. controls (232.04,55.6) and (230.81,57.11) .. (229.29,57.11) .. controls (227.77,57.11) and (226.53,55.6) .. (226.53,53.73) -- cycle ;
\draw [color={rgb, 255:red, 74; green, 144; blue, 226 }  ,draw opacity=1 ]   (79.11,33.44) -- (145.24,33.44) ;
\draw [color={rgb, 255:red, 74; green, 144; blue, 226 }  ,draw opacity=1 ]   (145.24,33.44) -- (150.74,45.1) ;
\draw [shift={(152.02,47.81)}, rotate = 244.77] [fill={rgb, 255:red, 74; green, 144; blue, 226 }  ,fill opacity=1 ][line width=0.08]  [draw opacity=0] (8.93,-4.29) -- (0,0) -- (8.93,4.29) -- cycle    ;
\draw [color={rgb, 255:red, 168; green, 168; blue, 168 }  ,draw opacity=1 ] [dash pattern={on 0.84pt off 2.51pt}]  (162.93,48.63) .. controls (172.88,63) and (175.44,64.82) .. (188.13,48.53) ;
\draw [shift={(189.33,46.97)}, rotate = 127.43] [color={rgb, 255:red, 168; green, 168; blue, 168 }  ,draw opacity=1 ][line width=0.75]    (10.93,-3.29) .. controls (6.95,-1.4) and (3.31,-0.3) .. (0,0) .. controls (3.31,0.3) and (6.95,1.4) .. (10.93,3.29)   ;
\draw [shift={(161.78,46.97)}, rotate = 55.46] [color={rgb, 255:red, 168; green, 168; blue, 168 }  ,draw opacity=1 ][line width=0.75]    (10.93,-3.29) .. controls (6.95,-1.4) and (3.31,-0.3) .. (0,0) .. controls (3.31,0.3) and (6.95,1.4) .. (10.93,3.29)   ;
\draw  [fill={rgb, 255:red, 245; green, 166; blue, 35 }  ,fill opacity=1 ] (13.33,33.94) -- (24.36,33.94) -- (24.36,61) -- (13.33,61) -- cycle ;
\draw  [fill={rgb, 255:red, 74; green, 144; blue, 226 }  ,fill opacity=1 ] (24.36,33.94) -- (35.38,33.94) -- (35.38,61) -- (24.36,61) -- cycle ;
\draw  [fill={rgb, 255:red, 245; green, 166; blue, 35 }  ,fill opacity=1 ] (35.38,33.94) -- (40.89,33.94) -- (40.89,61) -- (35.38,61) -- cycle ;
\draw  [fill={rgb, 255:red, 74; green, 144; blue, 226 }  ,fill opacity=1 ] (40.89,33.94) -- (62.93,33.94) -- (62.93,61) -- (40.89,61) -- cycle ;
\draw  [fill={rgb, 255:red, 245; green, 166; blue, 35 }  ,fill opacity=1 ] (62.93,33.94) -- (68.44,33.94) -- (68.44,61) -- (62.93,61) -- cycle ;
\draw  [fill={rgb, 255:red, 74; green, 144; blue, 226 }  ,fill opacity=1 ] (68.44,33.94) -- (96,33.94) -- (96,61) -- (68.44,61) -- cycle ;

\draw (161.6,3.38) node [anchor=north west][inner sep=0.75pt]  [font=\footnotesize] [align=left] {$\mathcal{D}_{\text{pool}}$};
\draw (95,90.15) node [anchor=north west][inner sep=0.75pt]  [font=\footnotesize] [align=left] {{Exploration}\\{Acq. function }};
\draw (197,90.15) node [anchor=north west][inner sep=0.75pt]  [font=\footnotesize] [align=left] {{Exploitation}\\{Acq. function }};
\draw (171,157) node [anchor=north west][inner sep=0.75pt]   [align=left] {$\mathcal{D}_q^*$};
\draw (12,18) node [anchor=north west][inner sep=0.75pt]  [font=\footnotesize] [align=left] {\textit{\textcolor[rgb]{0.29,0.56,0.89}{Annealing decision}}};

\end{tikzpicture}

%% file: figures/sensitivity-analysis/series.tex
\begin{tikzpicture}[baseline=(current bounding box.north)]

\definecolor{black3700}{RGB}{37,0,0}
\definecolor{black59155}{RGB}{59,15,5}
\definecolor{chocolate19911438}{RGB}{199,114,38}
\definecolor{coral25515351}{RGB}{255,153,51}
\definecolor{darkgrey176}{RGB}{176,176,176}
\definecolor{lightgrey204}{RGB}{204,204,204}
\definecolor{maroon873511}{RGB}{87,35,11}
\definecolor{saddlebrown1437424}{RGB}{143,74,24}

\begin{axis}[
width=0.5\textwidth,
height=0.26\textwidth,
legend cell align={left},
legend columns=1,
legend style={
  font=\scriptsize,
  fill opacity=0.8,
  draw opacity=1,
  text opacity=1,
  at={(0.97,0.03)},
  anchor=south east,
  draw=lightgrey204
},
tick align=outside,
tick pos=left,
x grid style={darkgrey176},
xlabel={Time (s)},
xmin=-7.87941014528275, xmax=185.498870108128,
xtick style={color=black},
y grid style={darkgrey176},
ylabel={Accuracy},
ymin=0.139716666666667, ymax=0.578616666666666,
ytick style={color=black}
]
\addplot [semithick, black3700, mark=square*, mark size=1, mark options={solid}]
table {%
0.910511684417725 0.167666666666667
1.31541805267334 0.206666666666667
1.68003423213959 0.26
2.08582587242126 0.29
2.45582504272461 0.302666666666667
2.86701714992523 0.318
3.24936461448669 0.337666666666667
3.65999960899353 0.362
4.04663710594177 0.371666666666667
4.4623095035553 0.377666666666667
4.85554647445679 0.386666666666667
5.27789399623871 0.404
5.67601594924927 0.404666666666667
6.11523239612579 0.421333333333333
6.58721210956573 0.42
7.79784216880798 0.42
9.15507957935333 0.420333333333333
9.98434607982635 0.432333333333333
11.1337760210037 0.454
12.3918306827545 0.454666666666667
13.2298126220703 0.454666666666667
14.4695143461227 0.452
15.7470429897308 0.462333333333333
16.8010927915573 0.468333333333333
18.1009786844253 0.477
19.2718079328537 0.477333333333333
20.7055781841278 0.482333333333333
21.9886633872986 0.481
23.2635971307754 0.487666666666667
24.5695767402649 0.493666666666667
25.9915138721466 0.495333333333333
27.4053747653961 0.501
28.810565495491 0.496
29.9034519910812 0.496666666666667
31.3049797296524 0.496333333333333
32.3695026636124 0.498333333333333
33.8287534952164 0.501
34.7837930679321 0.501666666666667
35.2895170211792 0.503333333333333
35.7851442575455 0.503
};
\addlegendentry{\(\displaystyle K\)-Centers \(\displaystyle \kappa=\)1}
\addplot [semithick, black59155, mark=*, mark size=1, mark options={solid}]
table {%
4.42232728004456 0.173333333333333
6.6688916683197 0.216333333333333
8.91857705116272 0.273333333333333
11.1716621160507 0.295
13.4186937093735 0.320333333333333
15.6791506290436 0.336333333333333
17.9369627475739 0.351666666666667
20.2027824640274 0.368
22.4814936161041 0.376
24.728497505188 0.39
27.26389067173 0.402666666666667
29.3713712692261 0.414333333333333
31.4794050931931 0.424
33.5892069578171 0.428666666666667
35.6947046518326 0.439
37.8039113283157 0.448333333333333
39.9134383916855 0.457
42.0355877161026 0.465666666666667
44.1506886482239 0.465
46.2693965435028 0.468
48.3920675754547 0.465666666666667
50.5204566717148 0.471
52.6475837945938 0.478333333333333
54.7792004585266 0.492333333333333
56.9117433547974 0.498666666666667
59.0490892887116 0.500333333333333
61.1837747097015 0.501666666666667
63.3187820196152 0.498333333333333
65.4629189491272 0.5
67.6059293031692 0.507333333333333
69.7506525993347 0.513666666666667
71.9007363796234 0.515666666666667
74.0567222356796 0.515666666666667
76.2107577562332 0.524666666666667
78.3697703838348 0.534
80.527764248848 0.537333333333333
82.6834218025208 0.533666666666667
84.8402768135071 0.532333333333333
86.9996365070343 0.53
89.1638789653778 0.533333333333333
};
\addlegendentry{Series aggregation \(\displaystyle \kappa=\)5}
\addplot [semithick, maroon873511, mark=*, mark size=1, mark options={solid}]
table {%
4.41302180290222 0.163333333333333
7.38232769966126 0.198333333333333
9.45365531444549 0.245
11.5301057100296 0.272666666666667
13.6053880214691 0.295333333333333
15.6866186141968 0.32
17.7817145824432 0.337333333333333
19.8786829471588 0.356
21.9738142251968 0.366333333333333
24.070276427269 0.371333333333333
26.1655028343201 0.383666666666667
28.269803404808 0.398333333333333
30.3696094751358 0.415666666666667
32.4730441808701 0.429
34.572802233696 0.434
36.679915356636 0.439333333333333
38.7877801656723 0.444666666666667
40.8950253486633 0.456666666666667
43.0029364109039 0.461333333333333
45.1143789291382 0.470333333333333
47.2261862516403 0.477
49.344468998909 0.477
51.4691572189331 0.479
53.6047163486481 0.478333333333333
55.7381696462631 0.490333333333333
57.8629997253418 0.493333333333333
59.9876212596893 0.495333333333333
62.1177579164505 0.498
64.2514871835709 0.499333333333333
66.388217496872 0.502666666666667
68.5293937683106 0.506
70.6680277109146 0.515
72.8067237138748 0.521
74.9497106790543 0.529333333333333
77.0893579959869 0.526666666666667
79.2440048217774 0.513
81.3925952196121 0.507333333333333
83.5415101051331 0.503666666666667
85.6905716180802 0.517666666666667
87.8488985061646 0.520333333333333
};
\addlegendentry{Series aggregation \(\displaystyle \kappa=\)10}
\addplot [semithick, saddlebrown1437424, mark=*, mark size=1, mark options={solid}]
table {%
4.49865684509277 0.162666666666667
6.88759911060333 0.206
9.09791531562805 0.256333333333333
11.4951500892639 0.284666666666667
13.7113705873489 0.306
16.1105007410049 0.326
18.7789921045303 0.343666666666667
22.9377926111221 0.36
25.1695396184921 0.371
27.493275642395 0.392333333333333
29.7153880357742 0.407
32.0392336130142 0.427333333333333
34.2710410118103 0.429333333333333
36.5948143959045 0.434666666666667
38.8333939790726 0.44
41.0621921300888 0.450666666666667
43.335276556015 0.466666666666667
48.147660946846 0.465
52.7117650985718 0.470666666666667
57.4303330659866 0.462
61.2920762062073 0.468666666666667
63.4561260223389 0.469333333333333
65.7208157777786 0.483666666666667
67.8949359178543 0.491333333333333
70.1518042325974 0.489333333333333
72.3243324518204 0.488
74.5889622449875 0.491333333333333
76.7680943250656 0.498
79.0405720949173 0.499666666666667
81.2268261909485 0.501666666666667
83.5108190536499 0.506
85.6981929302216 0.514666666666667
87.9840229511261 0.524333333333333
90.1743379592896 0.527
92.4562917232514 0.542666666666667
97.3994429826737 0.547666666666667
99.6876575469971 0.557
101.944322800636 0.544666666666667
104.235516309738 0.546666666666667
106.494113779068 0.551333333333333
};
\addlegendentry{Series aggregation \(\displaystyle \kappa=\)20}
\addplot [semithick, chocolate19911438, mark=*, mark size=1, mark options={solid}]
table {%
6.63571135997772 0.162
10.2481578111649 0.198
12.4424701213837 0.249666666666667
14.7129926681519 0.274666666666667
16.8620723247528 0.293333333333333
19.0920449018478 0.302
21.2374009370804 0.319
23.4728119850159 0.332666666666667
25.6175059556961 0.349
27.8630787372589 0.351666666666667
30.0156341552734 0.365666666666667
32.2627490997315 0.374333333333333
34.415514922142 0.389333333333333
36.6652056694031 0.395
38.8272651195526 0.405333333333333
41.0460265874863 0.409
43.2159811973572 0.414666666666667
45.4467797994614 0.421
47.6141149997711 0.428666666666667
49.8482013225555 0.435
52.0209517717362 0.438333333333333
54.3459460020065 0.449666666666667
56.5260633707047 0.465
58.8118313789368 0.473333333333333
61.0632655858994 0.472666666666667
63.3499336481094 0.470333333333333
65.6119514226913 0.476
67.9026073455811 0.479666666666667
70.1641115427017 0.480333333333333
72.4607042312622 0.487333333333333
74.7602598428726 0.494333333333333
77.0306514024735 0.507
79.3058485031128 0.502666666666667
81.5456853151321 0.507666666666667
83.8312920808792 0.508666666666667
86.1039387464523 0.516666666666667
88.3765567302704 0.513
90.7205852746964 0.512333333333333
93.0027680158615 0.508
95.3425960063934 0.511
};
\addlegendentry{Series aggregation \(\displaystyle \kappa=\)30}
\addplot [semithick, coral25515351, mark=square*, mark size=1, mark options={solid}]
table {%
9.75370509624481 0.159666666666667
13.1272212028503 0.204333333333333
16.4962359666824 0.259
19.8748531579971 0.289333333333333
23.2503029346466 0.314666666666667
26.6209710836411 0.334
29.9912873744965 0.352333333333333
33.3603084564209 0.362666666666667
36.7317168474197 0.372333333333333
40.1044851541519 0.378333333333333
43.4719086647034 0.391333333333333
46.8584807395935 0.405
50.2612021207809 0.423333333333333
53.6557584524155 0.426333333333333
57.0457684755325 0.429333333333333
60.4455624341965 0.435666666666667
63.8428739070892 0.445333333333333
67.2390361785889 0.455666666666667
70.726309132576 0.460333333333333
74.2343199491501 0.467333333333333
77.6366916179657 0.471666666666667
81.1292649507523 0.477333333333333
84.619752573967 0.488
88.1097082614899 0.499
91.6014092922211 0.508333333333333
95.091735458374 0.510666666666666
98.5827149152756 0.511666666666667
102.163251900673 0.519
108.090566396713 0.529
113.644144940376 0.539
119.583148550987 0.543666666666667
125.224665403366 0.544666666666667
130.904111790657 0.546
136.755769443512 0.541666666666667
142.168583774567 0.543333333333333
149.185871195793 0.538666666666667
157.244623398781 0.542333333333333
164.422576665878 0.54
170.459297156334 0.551
176.708948278427 0.558666666666666
};
\addlegendentry{BALD \(\displaystyle \kappa=\)40}
\end{axis}

\end{tikzpicture}

%% file: figures/sensitivity-analysis/annealing.tex
\begin{tikzpicture}

\definecolor{cyan}{RGB}{0,255,255}
\definecolor{darkgrey176}{RGB}{176,176,176}
\definecolor{darksalmon219164124}{RGB}{219,164,124}
\definecolor{indianred18312393}{RGB}{183,123,93}
\definecolor{lawngreen}{RGB}{124,252,0}
\definecolor{lightgrey204}{RGB}{204,204,204}
\definecolor{magenta}{RGB}{255,0,255}
\definecolor{maroon7600}{RGB}{76,0,0}
\definecolor{navajowhite255205155}{RGB}{255,205,155}
\definecolor{saddlebrown1124131}{RGB}{112,41,31}
\definecolor{sienna1478262}{RGB}{147,82,62}

\begin{axis}[
width=0.5\textwidth,
height=0.34\textwidth,
legend cell align={left},
legend style={
  font=\small,
  fill opacity=0.8,
  draw opacity=1,
  text opacity=1,
  at={(0.97,0.03)},
  anchor=south east,
  draw=lightgrey204
},
tick align=outside,
tick pos=left,
x grid style={darkgrey176},
xlabel={Time (s)},
xmin=-51.5472255860056, xmax=1260.38379649605,
xtick style={color=black},
y grid style={darkgrey176},
ylabel={Accuracy},
ymin=0.173833333333333, ymax=0.718071428571428,
ytick style={color=black}
]
\addplot [semithick, magenta, mark=*, mark size=1, mark options={solid}]
table {%
19.3474966117314 0.201904761904762
40.126496212823 0.288095238095238
62.1868880816868 0.38
85.4500777040209 0.448571428571429
110.193266493934 0.466666666666667
137.090776852199 0.499047619047619
164.682173456464 0.513333333333333
192.966957637242 0.547619047619048
223.178429807935 0.56
253.634071895054 0.575714285714286
285.933905158724 0.577142857142857
319.660281283515 0.585714285714286
354.602342128754 0.602380952380952
393.488925286702 0.62
433.345447369984 0.634285714285714
473.44507251467 0.638095238095238
515.801340716226 0.633809523809524
558.560657160623 0.637142857142857
604.286086116518 0.647619047619048
651.506904397692 0.658095238095238
700.771344866071 0.664285714285714
752.852122068405 0.657142857142857
805.441390003477 0.664285714285714
856.834691865104 0.665238095238095
911.442886488778 0.674285714285714
970.187760795866 0.679523809523809
1027.22279248919 0.688571428571429
1083.95032460349 0.690952380952381
1142.59402435166 0.68047619047619
1200.75056821959 0.675238095238095
};
\addlegendentry{BADGE}
\addplot [semithick, lawngreen, mark=*, mark size=1, mark options={solid}]
table {%
9.08155257361276 0.207142857142857
21.3024370329721 0.270952380952381
34.786141872406 0.361904761904762
48.8999797957284 0.417142857142857
61.9854281970433 0.461904761904762
76.2950547763279 0.490952380952381
96.0051958901542 0.514761904761905
121.96544442858 0.534761904761905
149.947229453496 0.558571428571429
176.109689167568 0.576666666666667
206.786550045013 0.593333333333333
241.928381102426 0.593333333333333
278.541273934501 0.606666666666667
323.600776059287 0.617142857142857
364.446847166334 0.627619047619048
348.247328874043 0.621428571428571
388.960169049672 0.619523809523809
432.202629464013 0.623333333333333
476.289370352881 0.632857142857143
522.133491492271 0.645238095238095
574.116273883411 0.658571428571429
623.277198045594 0.663333333333333
681.703188180923 0.665238095238095
737.011664857183 0.67
796.522250611441 0.674285714285714
856.35025300639 0.685714285714286
919.170227476529 0.681428571428571
982.780004259518 0.678571428571428
1045.4991042035 0.672380952380952
1111.07720900944 0.674761904761905
};
\addlegendentry{BALD}
\addplot [semithick, cyan, mark=*, mark size=1, mark options={solid}]
table {%
8.08600269045149 0.198571428571429
24.947053023747 0.269523809523809
39.3436234337943 0.364761904761905
58.9830656051636 0.429047619047619
83.9954511778695 0.473809523809524
102.630916867937 0.504285714285714
126.241513865335 0.534761904761905
151.014657429286 0.553333333333333
181.662245682308 0.55952380952381
213.999568871089 0.567619047619048
240.032372679029 0.585714285714286
264.258523396083 0.613809523809524
293.08728824343 0.624285714285714
323.17967401232 0.627142857142857
358.837492465973 0.620952380952381
398.842974594661 0.627619047619048
444.117510931832 0.636190476190476
483.23818901607 0.644761904761905
531.143505232675 0.65
577.277982643672 0.66
625.089121750423 0.663333333333333
673.502272537776 0.667619047619048
719.073296070099 0.671428571428572
758.604638576508 0.677619047619048
811.424848624638 0.680476190476191
870.258585521153 0.680476190476191
926.189272063119 0.682857142857143
986.856106689998 0.681428571428572
1050.79026562827 0.676190476190476
1106.42250101907 0.687142857142857
};
\addlegendentry{Random}
\addplot [semithick, maroon7600, mark=*, mark size=1, mark options={solid}]
table {%
18.4610336848668 0.21
37.7341129439218 0.284761904761905
58.8670665536608 0.372857142857143
81.2708873748779 0.42952380952381
105.521753447396 0.476666666666667
118.28471670832 0.50952380952381
130.904488870076 0.535714285714286
143.131806986673 0.557619047619048
156.93828357969 0.581428571428572
186.763630969184 0.598095238095238
217.529017175947 0.603333333333333
250.980264493397 0.616190476190476
284.662057774408 0.622857142857143
321.047919613974 0.637619047619048
341.903102295739 0.646666666666667
366.20338337762 0.654285714285714
390.328171559743 0.654761904761905
413.171010562352 0.655238095238095
454.855103322438 0.664285714285714
496.767604555402 0.671904761904762
542.231288160597 0.68047619047619
588.089348077774 0.68047619047619
635.292360986982 0.680952380952381
670.134472302028 0.682380952380952
706.294300692422 0.681428571428571
745.814756120954 0.674761904761905
785.152460643223 0.674285714285714
841.372833933149 0.677142857142857
897.701048510415 0.688571428571429
956.057620321001 0.688571428571429
};
\addlegendentry{Annealing \(\displaystyle r=1\)}
\addplot [semithick, saddlebrown1124131, mark=*, mark size=1, mark options={solid}]
table {%
21.7673592908042 0.204285714285714
44.4154140949249 0.263809523809524
68.5906021935599 0.348095238095238
93.2918903827667 0.393809523809524
120.136227743966 0.445714285714286
131.581701346806 0.476190476190476
143.798855781555 0.511428571428571
157.443500484739 0.527142857142857
171.964728934424 0.53952380952381
205.456807545253 0.553333333333333
240.910239049367 0.564285714285714
277.774851151875 0.577619047619048
318.791544539588 0.588095238095238
361.444736071995 0.600952380952381
386.476619720459 0.611428571428571
412.57468192918 0.626666666666667
436.990205560412 0.633809523809524
465.169434853963 0.636666666666667
510.208900077002 0.637619047619048
555.508920669556 0.640952380952381
604.411856651306 0.657619047619048
653.908528498241 0.659523809523809
703.306121179036 0.666666666666667
742.9137403965 0.678095238095238
781.700627531324 0.687142857142857
826.039122172764 0.692857142857143
868.109372207097 0.684285714285714
928.634083509445 0.683333333333333
985.514206171036 0.684761904761905
1046.31693754877 0.688095238095238
};
\addlegendentry{Annealing \(\displaystyle r=1.2\)}
\addplot [semithick, sienna1478262, mark=*, mark size=1, mark options={solid}]
table {%
21.1119684491839 0.200952380952381
43.2066585677011 0.275714285714286
66.9400496142251 0.368095238095238
91.3516089235033 0.425714285714286
117.764019727707 0.457142857142857
128.938669272832 0.461904761904762
142.331241062709 0.479047619047619
155.380134071623 0.496666666666667
168.960858004434 0.524285714285714
201.507159369332 0.543809523809524
229.843846627644 0.566666666666667
263.898507424763 0.584761904761905
294.361646141325 0.603333333333333
334.285029172897 0.612380952380953
354.902704954147 0.619047619047619
375.489733764104 0.626190476190476
397.344496011734 0.630952380952381
420.314835344042 0.633809523809524
460.218597446169 0.634761904761905
500.518215179443 0.645714285714286
544.197793688093 0.668095238095238
587.855504580906 0.678571428571428
636.441871915545 0.678571428571428
670.345274482455 0.681428571428571
706.716113703592 0.679523809523809
745.575700623648 0.686666666666667
787.475937673024 0.682380952380952
842.362033741815 0.69
898.548347813742 0.69047619047619
956.766496624265 0.693333333333333
};
\addlegendentry{Annealing \(\displaystyle r=1.4\)}
\addplot [semithick, indianred18312393, mark=*, mark size=1, mark options={solid}]
table {%
18.0926514693669 0.214285714285714
37.2300942965916 0.291428571428571
58.128935779844 0.39
81.154541015625 0.441428571428571
105.270558050701 0.480952380952381
115.281685556684 0.497142857142857
126.820657627923 0.523809523809524
139.224925892694 0.545238095238095
152.467565604619 0.570952380952381
183.809438262667 0.589047619047619
214.863675866808 0.605714285714286
247.18473468508 0.618095238095238
278.642426899501 0.624761904761905
314.539873702186 0.631428571428572
335.179357835225 0.644761904761905
354.720580305372 0.653333333333333
379.043432712555 0.653333333333333
402.836670024054 0.648571428571429
444.177111966269 0.65
484.197566475187 0.649047619047619
526.411400113787 0.651428571428572
569.323821408408 0.658095238095238
615.300741400038 0.664285714285714
644.388138805117 0.660476190476191
676.554881061826 0.660952380952381
711.444766146796 0.673333333333333
747.810654674258 0.687619047619048
802.862159115928 0.687142857142857
856.914849042892 0.684761904761905
913.451232671738 0.681904761904762
};
\addlegendentry{Annealing \(\displaystyle r=1.6\)}
\addplot [semithick, darksalmon219164124, mark=*, mark size=1, mark options={solid}]
table {%
20.8311923571995 0.205238095238095
42.6579342229026 0.276666666666667
65.8982989447457 0.368095238095238
89.7659911087581 0.424761904761905
115.155653033938 0.462857142857143
124.633190529687 0.488571428571428
135.593136344637 0.504761904761905
148.825550760542 0.524285714285714
162.533949920109 0.550952380952381
193.922156947 0.576190476190476
226.542418718338 0.602380952380952
261.981993504933 0.596190476190476
299.254523243223 0.610952380952381
336.951202835356 0.615714285714286
359.6050929342 0.641428571428571
384.121585914067 0.65
409.859010764531 0.648571428571428
434.548638684409 0.653333333333333
478.801089763641 0.659523809523809
524.522521972656 0.676190476190476
567.527487788882 0.682380952380952
612.575055224555 0.682380952380952
664.780331918171 0.685714285714286
702.620216539928 0.69
744.704993929182 0.684761904761905
789.022125550679 0.67952380952381
834.071738617761 0.682857142857143
887.571124247142 0.686666666666667
940.623415810721 0.685238095238095
998.257230724607 0.68
};
\addlegendentry{Annealing \(\displaystyle r=1.8\)}
\addplot [semithick, navajowhite255205155, mark=*, mark size=1, mark options={solid}]
table {%
17.7113300391606 0.208571428571429
37.5063966342381 0.281428571428571
57.192437171936 0.375714285714286
78.8586621965681 0.425714285714286
102.173691068377 0.465714285714286
111.103248085294 0.48
120.40847161838 0.507142857142857
132.317937782833 0.524761904761905
146.84132875715 0.555238095238095
176.024268967765 0.571904761904762
205.020696163177 0.592380952380952
238.253882169724 0.599047619047619
272.17453953198 0.608571428571429
309.640898908888 0.609523809523809
326.703542947769 0.62
346.857969624656 0.632857142857143
371.433653354645 0.644761904761905
392.845731224333 0.656190476190476
417.16330242157 0.665714285714286
445.775942121233 0.67
470.05293733733 0.667142857142857
500.554322821753 0.664285714285714
526.979045425143 0.664285714285714
571.090358359473 0.67
618.069856848036 0.666666666666667
666.830654178347 0.67
717.442540952138 0.670476190476191
766.749836921692 0.675238095238095
804.790149177824 0.676666666666667
846.534731149673 0.685238095238096
};
\addlegendentry{Annealing \(\displaystyle r=2\)}
\end{axis}

\end{tikzpicture}